\documentclass[conference]{IEEEtran}
\IEEEoverridecommandlockouts
\usepackage{cite}
\usepackage{amsmath,amssymb,amsfonts}
\usepackage{algorithmic}
\usepackage{graphicx}
\usepackage{textcomp}
\usepackage{xcolor}
\usepackage{hyperref}
\usepackage{multicol}
\usepackage{subfigure}
\usepackage{caption}
\usepackage{booktabs}
\usepackage{ctable}
\RequirePackage{multirow}
\def\BibTeX{{\rm B\kern-.05em{\sc i\kern-.025em b}\kern-.08em
    T\kern-.1667em\lower.7ex\hbox{E}\kern-.125emX}}
\begin{document}

\title{Predicting Visual Attention and Distraction During Visual Search Using Convolutional Neural Networks\\
{\footnotesize \textsuperscript{*}This is a pre-print manuscript currently under review in Journal of Vision.}
}

\author{\IEEEauthorblockN{Manoosh Samiei}
\IEEEauthorblockA{\textit{Electrical and Computer Engineering Department} \\
\textit{McGill University}\\
Montreal, Quebec, Canada \\
manoosh.samiei@mail.mcgill.ca}
\and
\IEEEauthorblockN{James J. Clark}
\IEEEauthorblockA{\textit{Electrical and Computer Engineering Department} \\
\textit{McGill University}\\
Montreal, Quebec, Canada \\
james.clark1@mcgill.ca}
}

\maketitle

\begin{abstract}
Most studies in computational modeling of visual attention encompass task-free observation of images. Free-viewing saliency considers limited scenarios of daily life. Most visual activities are goal-oriented and demand a great amount of top-down attention control. Visual search task demands more top-down control of attention, compared to free-viewing. In this paper, we present two approaches to model visual attention and distraction of observers during visual search. Our first approach adapts a light-weight free-viewing saliency model to predict eye fixation density(probability) maps of human observers over pixels of search images, using a two-stream convolutional encoder-decoder network, trained and evaluated on COCO-Search18 dataset. This method predicts which locations are more distracting when searching for a particular target. Our network achieves good results on standard saliency metrics (\textit{AUC-Judd}=0.95, \textit{AUC-Borji}=0.85, \textit{sAUC}=0.84, \textit{NSS}=4.64, \textit{KLD}=0.93, \textit{CC}=0.72, \textit{SIM}=0.54, and \textit{IG}=2.59). Our second approach is object-based and predicts the distractor and target objects during visual search. Distractors are all objects except the target that observers fixate on during search. This method uses a Mask-RCNN segmentation network pre-trained on MS-COCO and fine-tuned on COCO-Search18 dataset. We release our segmentation annotations of targets and distractors in COCO-Search18 for three target categories: bottle, bowl, and car. The average scores over the three categories are: \textit{F1-score}=0.64, \textit{MAP\textsubscript{0.5}}=0.57, \textit{MAR\textsubscript{0.5}}=0.73. Our implementation code in Tensorflow is publicly available at \url{https://github.com/ManooshSamiei/Distraction-Visual-Search}.
\end{abstract}


\section{Introduction}

How do we find our keys in a messy room among a pile of papers and clothes? How do we find our favorite jam among other products on the supermarket shelves? How do we identify our friend among a huge crowd of students in a school yard? These tasks seem trivial for humans but if we try to teach a machine to perform the same tasks and achieve human-level performance, they seem challenging. Humans have evolved with an incredible capacity to scan their visual environment efficiently for finding food and avoiding potential danger. Understanding the underlying strategies that human brain adopts during visual search is important both from a modeling perspective and also for behavioral studies. 

One of the key components of the human visual perception system that facilitates search is selective attention. Selective attention allows organisms to direct their gaze toward their object of interest to attend to only small parts of their visual environment that are more important to them, reducing the computational demand of vision. Moreover, humans have evolved with a foveated visual system, meaning that only \texttildelow 1 degree of visual angle around the fovea (center of the retina) has a high visual acuity due to higher density of receptors. Selective attention guides human eye movements to place a high-resolution image of their visual target on their fovea for a detailed perception. Among different types of eye movements, saccades and their associated fixations, play the most important roles in our perception. During a saccade, the gaze moves rapidly from one location to another. Hence, the image on the fovea during a saccade is of low quality. The perception of the environment is therefore done during eye fixations, where the gaze is maintained for a period of time on a target location between the saccades. Studying eye fixations during visual search can therefore provide key information regarding the visual attention and search policy of human brain. 

It is widely believed that visual attention is driven by two main components: bottom-up, using low-level stimuli-based features,  and top-down, using high-level goal-directed features. An example of bottom-up attention is the pop-out effect when a red object on a green background attracts observers' attention due to its high visibility. On the other hand, top-down attention takes part in any visual task associated with a goal, such as in driving. During driving, a driver regularly scans the scene for other cars, traffic signs, pedestrians, lane boundaries, and other objects important for the driving decisions. In general, bottom-up and top-down attention are combined in the control of gaze. Some studies \cite{article} \cite{article_l} suggest that as the computational demands (perceptual load) of a task increases, top-down attention dominates the low-level bottom-up attention. 

An example of a low-load task that is extensively studied in saliency research is free-viewing. In a free-viewing task, observers are looking at a set of images without any particular task and their eye fixations are recorded and modeled. This sort of task is mostly associated with bottom-up stimuli-driven attention. Free-viewing saliency encompasses a limited number of humans' daily activities, which are mostly goal-oriented and in need of top-down attention control. Most research on understanding the high-level goal-oriented element of attention has been done in the context of visual search. Visual search is a relatively simple task which involves more goal-directed attention compared to free-viewing. 

In this study, we present two approaches to model humans’ visual attention and distraction behavior during visual search. In our modelings, we consider any fixation on the non-target objects as a distraction, and our methods predict which regions or objects in an image are more distracting when searching for a particular target. Our first model predicts fixation density maps during visual search, given visual information about the search target. The model generates the probability of eye fixation over pixels of the search image, using a two-stream encoder-decoder network similar to MSI-Net \cite{KRONER2020261} with a VGG16 feature extraction backbone, initialized with pre-trained weights on ImageNet dataset for object detection task. One stream of our network receives a sample image of the search target from a specific target category, and the other receives the search image containing the target. The extracted features of the two streams are then convolved and passed through a decoder to generate a target-specific fixation density map for the given search image. This method predicts distracting regions at pixel-level. Our second method detects distractors and targets in search images at an object-level. We use a Mask-RCNN network with Resnet101 backbone, initialized with pre-trained weights on the MS-COCO dataset for object detection, to segment distractors and targets. The COCOSearch18 dataset is our main dataset for training/fine-tuning and testing our networks in both methods. Our segmentation annotations for targets and distractors of COCOSearch18 dataset for three target categories, namely bottle, bowl, and car are accessible at our public \href{https://github.com/ManooshSamiei/Distraction-Visual-Search/blob/main/target-distractor_segmentation/README.md}{GitHub} repository.

Predicting the distraction effects during search can be applied for commercial purposes such as improving the visibility of products on a super market shelf, or facilitating the usage of a web-page through minimizing distracting patterns. It can also be used for enhancing image/video quality by removing distractors that draw observers attention away from the main subject of the content. Another application could be in improving object detection and segmentation algorithms by introducing region proposal strategies that employ human attention to propose regions in the images that are likely to contain a target object.

\section{Related Work}

Rosenholtz \cite{ROSENHOLTZ19993157} \cite{10.1167/4.3.9} designed a mathematical model of visual search, in which the saliency of a target item is defined as the distance between distractors' features and target features. Navalpakkam and Itti \cite{integratedtopbottom} computed the optimal top-down weights to maximize the target's salience relative to the distractors by maximizing the signal-to-noise ratio (\textit{SNR}) of the target versus distractors. Torralba and Oliva et al. \cite{cotextualguidance} proposed a Bayesian framework for visual search tasks which includes both target features and contextual information of the scene in a probabilistic way. Zhang et al. proposed another Bayesian model of visual search called `SUN' (saliency using natural statistics) \cite{doi:10.1080/13506280902771138}. This model contains a bottom-up saliency component (similar to \cite{cotextualguidance}) and a top-down component that guides attention to the areas of the scene likely to be the target solely based on appearance. This goal is achieved by maximizing the point-wise mutual information between features and the target class. Ehinger et al. \cite{Ehinger2009ModellingSF} used the Bayesian framework of Torralba et al. to explain the eye movements in searching for people in a database of about 912 natural scenes. They computed the saliency maps predicted by each component of the framework, namely bottom-up saliency, gist, and target features; and then combined them by multiplying the weighted saliency maps. 

One of the first biologically inspired computational models that uses convolutional neural networks to model human visual search behavior is Invariant Visual Search Network (IVSN) \cite{FindWaldo} proposed by Zhang et al. This network locates targets through sequential fixations without exhaustive search.  The model consists of two separate streams of feed-forward convolutional neural network (VGG16) pre-trained for object recognition task on ImageNet dataset. One stream extracts features from a sample object from the target category (if the target object is a bottle, then a different bottle is presented to this stream of network) and another stream extracts features from the entire search image. The features of a sample target image are convolved with the feature map of a search image, and an attention map is generated. This process is an imitation of pre-frontal cortex top-down modulation that contains the task-dependent information about a target. Fixation locations are generated in the descending order of maximum locations in the attention map using a winner-take-all mechanism (WTA). If the target is not found at the current fixation, inhibition of return is applied (IOR) and the next maximum location is selected. This process is repeated until the target is found. In our proposed model we use a similar two-stream setting but in contrast to IVSN that uses pretrained convolutional neural network on ImageNet without any further training, we train the whole model end-to-end on human eye fixation maps to obtain a more accurate predictor of human visual search fixation locations. Another key difference of our model with IVSN, is our focus on spatial saliency rather than the spatio-temporal modeling of scanpaths. One shortcoming of IVSN is that its generated scanpaths are different from human scanpaths in that they can move between far parts of the image (i.e. going from bottom left corner to the top right corner); while, humans tend to move their eyes more smoothly. One key advantage of this model is its zero-shot feature meaning that it can generalize to novel objects without any training.

Numerous studies such as  \cite{10.1145/3313831.3376870}, \cite{JOKINEN2020102376}, \cite{431eb1481a9543249c08f942e64ec2b5}, \cite{10.1145/2207676.2208414}, \cite{doi:10.1080/07370020701638806}, and \cite{10.1145/3241381} present computational models of visual search on graphical layouts and web pages. \cite{10.1145/3173574.3173862} designed a model for grid tasks on touchscreen mobile devices by combining traditional analytical methods and data-driven machine learning approaches. Also, \cite{10.1145/2556288.2557093}, \cite{10.1145/2702123.2702483} and \cite{10.1145/1240624.1240723} provide models for menu search and performance.

Zelinsky et al.  \cite{9025609} proposed a model that predicts the scan-path (i.e. sequence of saccades and fixations) of observers while searching for either a clock or a microwave in a set of images which contain the natural context of these objects. They presented two models, one CNN-based and the other RNN-based. In their CNN-based method, they used cumulative foveated images, in which the information accumulates over fixations by progressively de-blurring a blurred foveated image based on high-resolution information obtained at each new fixation, similar to humans foveated visual system. Then they trained a convolutional neural network to input a cumulative foveated image based on a given fixation in a scanpath and output the location of the next fixation. The CNN is based on ResNet-50 which outputs a location probability map via a softmax layer. Finally, a WTA network chooses the location in output image with the highest probability as the next predicted fixation. Their second method involves training three types of recurrent neural networks (simple RNN, LSTM, and GRU), which is suitable for sequential data modeling. In their recurrent methods, they first extract features from images using ResNet-50, pre-trained on ImageNet, and then use this feature map to predict the next fixation given the previous fixation location. Among the two methods, RNN-based models outperformed the CNN model. One drawback of their methods is that a separate model should be trained for each target category, i.e. one model for microwave search and another for clock search. 

In their later work \cite{Goal-Directed-scan}, the same authors proposed a model to predict human scan-paths during visual search using inverse reinforcement learning (IRL). They attempt to learn the reward function and policy used by humans during visual search using adversarial training. In this formulation, reward and policy are considered as part of the discriminator and generator, respectively. The discriminator assigns high reward to a human-like behavior and low reward to a non-human behavior, where behavior is represented as state-action pairs. The generator/policy is optimized using a reinforcement learning algorithm, called GAIL (generative adversarial imitation learning) to get higher rewards by generating more human-like scan-paths. For the state representation they used a method called `dynamic contextual belief maps of object location'. Similar to their previous work, they implement the effect of fovea in their state representation, which they termed retina-transformed image. However, this time instead of a progressive blurring on the input image, for each fixation they placed a high-resolution local patch of the image around the fixated location and the blurred representation of the rest of the image as the peripheral input. Contextual belief maps are then created to account for the hypothesis that humans parse a scene into objects and backgrounds to make a belief map of the target's location, which then guides their eye movements. To create contextual belief maps, authors create panoptic segmentation of the scene, generating masks for 80 object and 54 background classes, and then grouping all mask instances belonging to the same category to create a single mask per category. These belief maps are generated for both high-resolution input image and low-resolution image. Then at each fixation, the state is updated by replacing the low resolution belief maps with the corresponding high-resolution map computed at the fixation location. To account for the scan-path dependence on the search target (task), authors concatenate a one-hot encoded task vector to the belief maps. To train and evaluate their model, authors created a large-scale dataset named COCO-Search18 \cite{cocos} including search fixations of 10 people viewing 6202 images while searching for each of 18 target-object categories. This is the dataset that we also use in this study to train and evaluate our proposed model. The authors further compare their model with 5 other models on scan-path evaluation metrics. They conclude that the IRL algorithm outperforms the other methods on all metrics. They also show that the reward maps recovered by the IRL model depend greatly on the category of the search target. Another advantage of this method is that only one model is trained and used for all target object categories. 

Chen et al. created a human eye-tracking dataset for Visual Question Answering (VQA) tasks \cite{chen2020air}. During the experiments, participants were asked several questions from an image such as the color of a particular object or the spatial relationship between two objects, and their eye movements were recorded. Visual question answering tasks include visual search as part of the process, but they also demand visual reasoning which makes them more complicated compared to visual search. 

There are several studies such as \cite{HAJIABOLHASSANI2014127} \cite{10.1167/14.3.29} \cite{BOISVERT2016653} that model the reverse operation. They attempt to infer the task of observers from their eye tracking data. Haji-Abolhassani and Clark \cite{HAJIABOLHASSANI2014127} used Hidden Markov Models to infer 4 types of tasks from observers' eye tracking data. The 4 tasks were: memorizing the picture, determining the decade in which the picture was taken, determining how well the people in the picture know each other, and determining the wealth of the people in the picture. 

Research on inferring task from eye tracking data validates the dependency between observers' scan-path patterns and their visual task. In \cite{HAJIABOLHASSANI2014127} the authors suggest that observers fixated on faces when they were asked about the decade that the picture was taken, while they fixated on inanimate objects for estimating the wealth of the people. All of these distinct task-driven features in eye movements can be used to study how human's visual attention works. 

One similar research article to our target-distractor segmentation method is the work of Fried et. al \cite{7298779}. The authors of this paper propose a method to predict distractors, which are defined as `the regions of an image that draw attention away from the main subjects and reduce the overall image quality.' They assign to each segmented region of an image, which is obtained using multi-scale combinatorial grouping (MCG) \cite{6909443}, a distraction score and remove the most distracting regions. The ground truth distraction score for each segment is calculated by taking the average distraction score of each pixel over all pixels in that segmented region. The distraction score of each pixel is computed as how many human annotators have considered that pixel as distracting. Features are manually extracted for each segment, and LASSO algorithm learns the mapping between the extracted features and the distraction score of that segment. Later, in 2018 \cite{8247231} uses a convolutional encoder-decoder architecture, called SegNet \cite{Badrinarayanan2015SegNetAD}, to predict a distractor map in each video frame. Both of these articles detect distractors in free-viewing condition, while our second approach predicts target and distractor segmentation during visual search.

\section{Dataset}

We used COCO-Search18 fixation dataset \cite{cocos} to train and evaluate our models. COCO-Search18 is a fixation dataset containing the fixation locations of 10 observers searching for each of the 18 object categories in 6202 images that are borrowed from COCO2014 \cite{coco2014} dataset. Among these 6202 images, 3101 images contain the target object (used for target-present trials) and 3101 do not contain the target (used for target-absent trials). In our modelings, we only use 3101 target-present images and their corresponding fixation data. The creators of the dataset, started each trial by showing a fixation dot at the center of the screen. Participants started a trial by pressing a button on a game-pad controller while looking at the fixation dot. An image of a scene was displayed and participant's task was to quickly decide if the target is present in the image or not, by responding `yes' or `no' using the  right or left triggers of a game-pad controller. The 18 categories of target objects are composed of bottle, bowl, car, chair, clock, cup, fork, keyboard, knife, laptop, microwave, mouse, oven, potted plant, sink, stop sign, toilet, and TV; all of which appear in their natural context. For instance, cars appear on the streets or other outdoor scenes, laptops appear on desks/tables at the offices, and microwaves appear in the kitchen environment. All images in the dataset were resized to 1680 × 1050 with zero padding and aspect ratio kept. The fixation data are presented in json files. The json files also contain the fixations' time duration and participants' reaction time. The dataset is accessible at \url{https://saliency.tuebingen.ai/datasets/COCO-Search18/}.

\section{Method 1: Predicting Salience During Search}
\label{sec:method}

We propose a double-input neural network, which given a search image and a target image, predicts the fixation density map of human observers searching for that target category in that image. A fixation density map (FDM) contains the probability $p(x,y|I)$ of observing a fixation at a given pixel in a given image. These maps represent the locations in an image that are likely to be considered salient by an observer while searching for a specific target category. In some research papers fixation density maps are referred to as saliency maps. However, we avoid this terminology as suggested by Kummerer et al. \cite{Kummerer_2018_ECCV}; they propose that a saliency map should be defined as a metric-specific prediction derived from the model fixation density, in order to generate the highest performance of the model for each saliency metric (such as \textit{IG}, \textit{NSS}, \textit{CC}, etc.).

\subsection{Model Architecture}

\begin{figure}[!htbp]
\centering
\includegraphics[scale = 0.35]{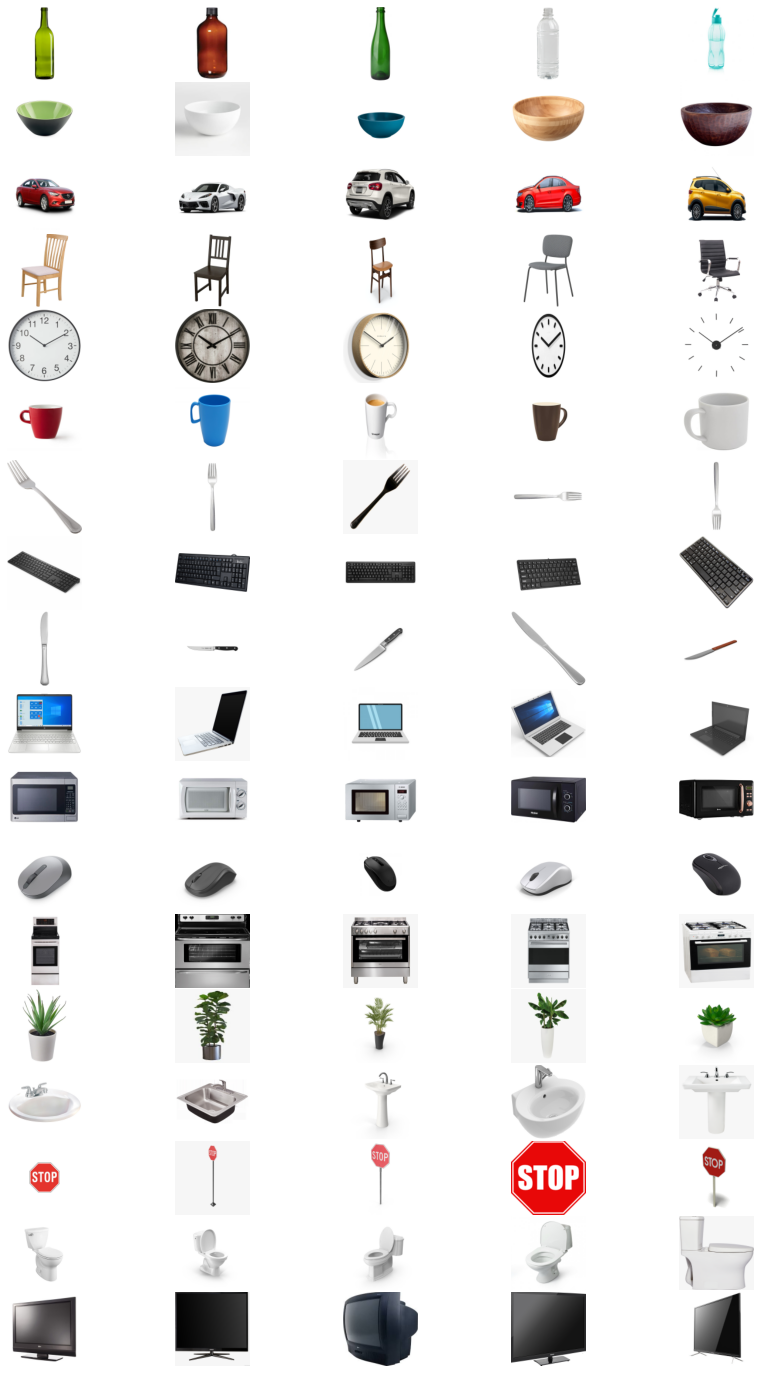}
\caption{Sample target images for the 18 object search categories. For instance, when we aim to generate fixation density map of searching for a TV, we randomly choose one of the five TV sample images presented in this figure, and feed it to the target stream of our network. All sample targets are resized to 64 x 64 dimension before entering the target stream.}
\label{fig:targets}
\end{figure}

In our modeling, we employ the light-weight convolutional encoder-decoder architecture of MSI-Net (Multi-scale Information Network) free-viewing saliency model \cite{KRONER2020261}. Our network has an additional encoder stream that receives a sample target image from the search target category. The sample target fed into the target stream is not the same as the target object in the search image. At each epoch, we randomly choose one sample target image from the five samples that we have for each of the 18 target categories and feed it to the target encoder stream. These sample target objects can be seen in figure \ref{fig:targets}.

Thus, one encoder stream of our network extracts features from the search image and the second encoder stream extracts features from a sample image of the target category. These two streams are identical and their parameters are shared. Each encoder stream is composed of two parts. The first part contains a VGG16 convolutional neural network \cite{brusilovsky:simonyan2014very} pretrained on ImageNet dataset \cite{ILSVRC15}, and the second part is an `ASPP' \cite{ASPP} module. VGG16 is composed of 5 blocks of convolutional layers with 3x3 kernels followed by pooling layers, and three final fully-connected layers. For the purpose of feature extraction, we remove the last fully-connected layers. All layers use zero-padding to keep the width and height
of their input tensors unchanged. Hence, only pooling layers cause reduction in dimensions. As in MSI-Net, we remove the strides of the last two pooling layers in VGG16 to keep the width and height of the features as $1/8$ of the input image. To be able to use the pretrained weights on Imagenet and match the same number of parameters, we use a dilation rate of 2 in the last three convolutional layers. To obtain a multi-scale feature representation, we concatenate the output of the three last pooling layers, all of which have an output dimension of $1/8$ of the input image. The pooling layer in block three has 256 output channels and pooling layers of blocks four and five each has 512 channels. Hence the concatenated features tensor has 1280 output channels. This tensor is then sent to the second parts of the feature extraction block called `Atrous Spatial Pyramid Pooling' abbreviated as `ASPP'. 

\begin{figure}[h]
\centering
\includegraphics[scale = 0.29]{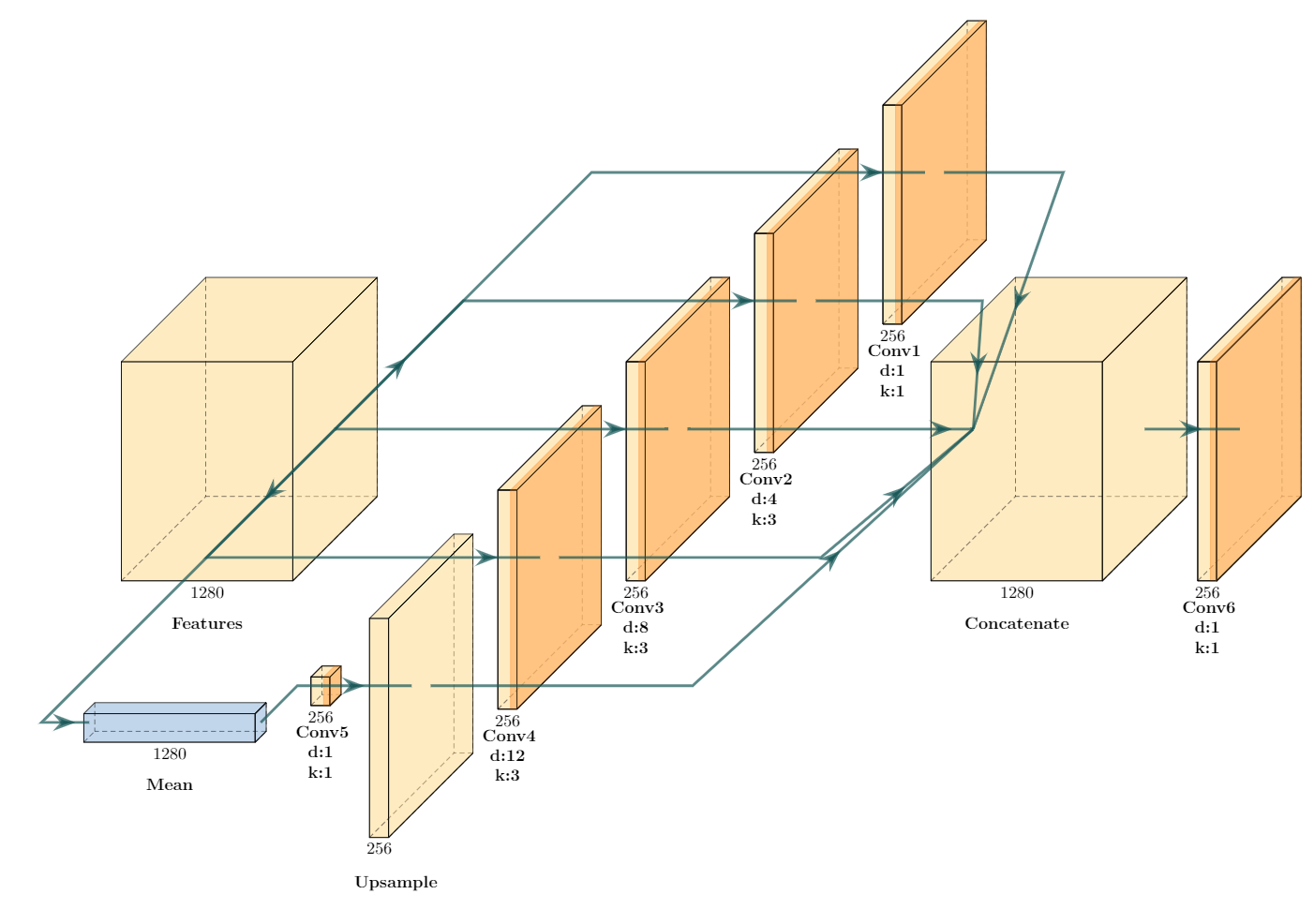}
\caption{ASPP Architecture.}
\label{fig:ASPP}
\end{figure}

ASPP is a semantic segmentation module that is composed of convolutional layers with different dilation rates (`Atrous' convolution is another term for dilated convolution). This causes the layers to use filters with different effective fields of view, that capture objects and contextual information at different scales. Same as MSI-Net, the ASPP architecture used in our network is composed of six convolutional layers. Four convolutional layers receive the extracted feature tensor (with 1280 channels) directly as their input. One of them has 1 × 1 kernels and only perform point-wise non-linearity without learning spatial relationships. The other three have 3 × 3 kernels with 4, 8, or 12 dilation rate, which allows extracting multi-scale dependencies from the feature tensor. The fifth convolution layer receives the mean of the features across spatial dimensions, which is believed to contain global contextual information of the image. This convolution then applies point-wise non-linearity with 1 × 1 kernels to the mean tensor. The output of this convolution is then upsampled to match the spatial dimension of the original feature tensor. Finally, the output of all these five convolutions are concatenated to form a single tensor. As the output of each convolution has 256 channels, the concatenated tensor yields 1280 channels. The concatenated tensor is then passed through another convolutional layer with point-wise 1 × 1 kernels and ReLU non-linearity. The output of this layer yields the final multi-scale feature tensor with 256 channels. The architecture of ASPP can be seen in image \ref{fig:ASPP}. 

\begin{figure*}[h]
\centering
\includegraphics[scale = 0.35]{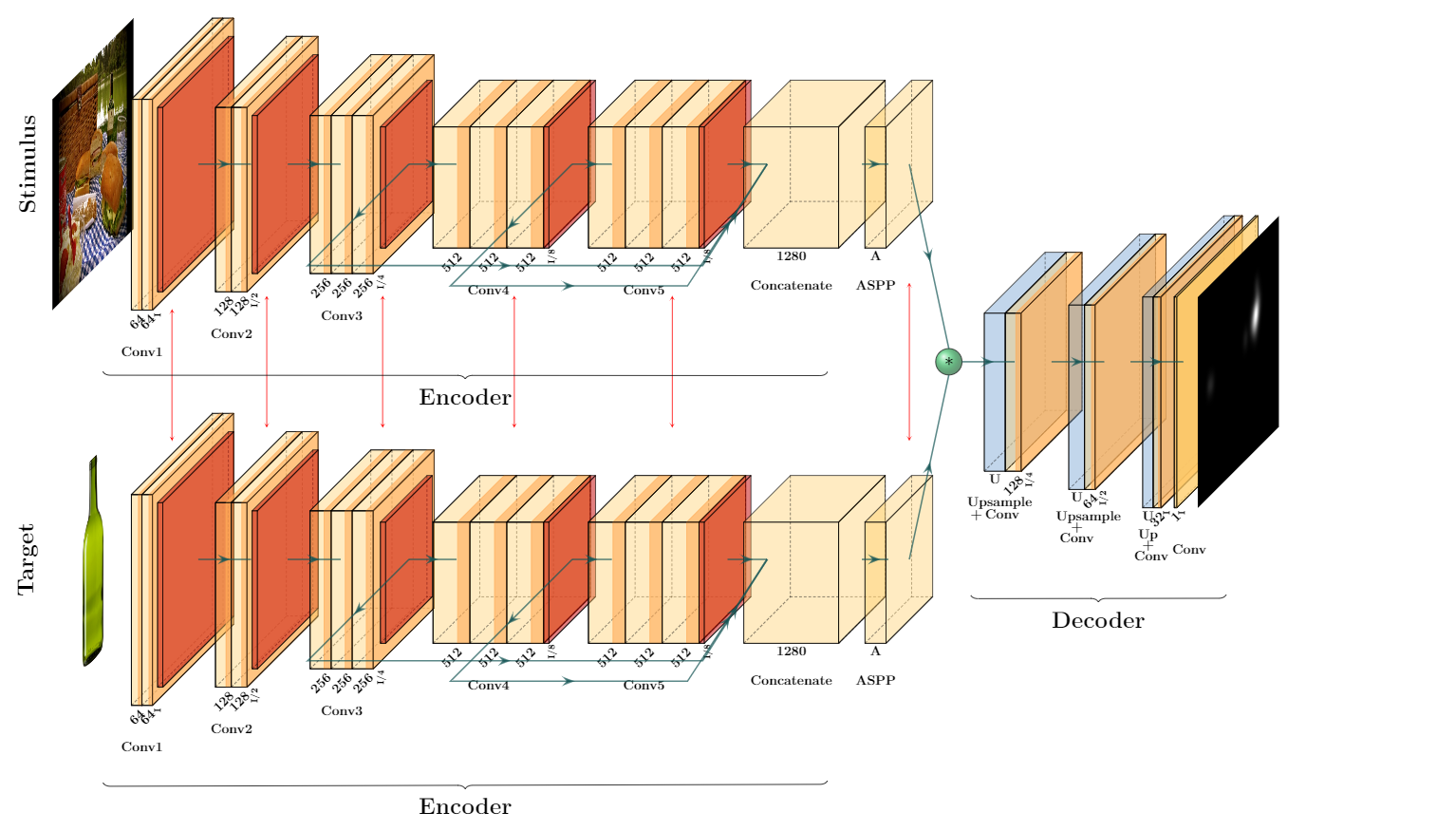}
\caption{Network architecture. The red arrows indicate that the weights of the convolution layers in encoder and ASPP modules are shared between the two streams.}
\label{fig:network}
\end{figure*}

Both the target and stimulus streams use the same encoder and ASPP architecture to extract features from the sample target image and search image. Then extracted features from the target and search image are convolved together. This convolution is performed using a convolutional layer with the target feature tensor treated as the filter and the search image feature tensor as the input to the layer, without any non-linear activation function. The output of this convolution has 256 channels, which is then passed through a decoder to retrieve the spatial dimension of the input image. Decoder is composed of three bi-linear up-sampling blocks, followed by 3 × 3 convolution layers to avoid checkerboard artifacts. There are four convolution layers in the decoder with 128, 64, 32, and 1 filters. The first three convolutions that come immediately after up-sampling layers are followed by ReLU non-linearity; while, the last convolution layer that generates the output FDM is not modified by a ReLU. The output values are then normalized between 0 and 1, such that all values sum to one. This modification converts the generated map to a probabilistic density map. A visualization of our network is presented in figure \ref{fig:network}.     

\subsection{Data Preprocessing}

\begin{figure}[h]
    \centering
    \subfigure[]{
    \includegraphics[width=0.2\textwidth]{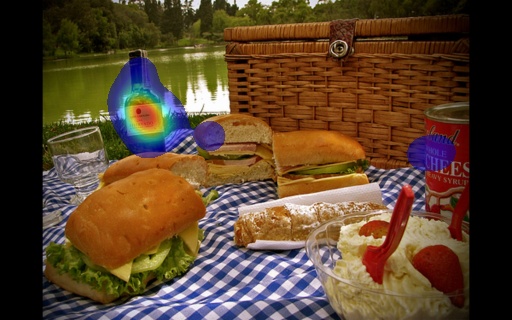}}
   \subfigure[]{
    \includegraphics[width=0.2\textwidth]{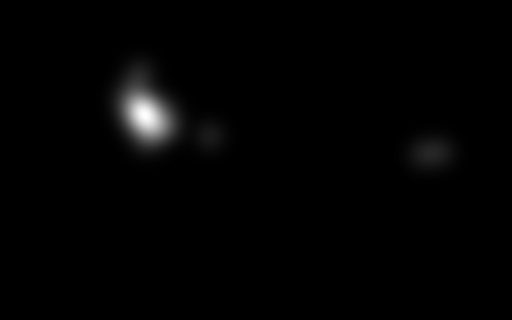}}
\caption{(a) Gaze heat-map overlaid on search image. (b) Ground truth fixation density map.}
\label{fig:maps}
\end{figure}

We created the fixation density maps for each image by blurring each fixation point with a Gaussian kernel centered at that point with a standard deviation of 11. We sum these Gaussians on the image plane, and then normalize the map by dividing it by the sum of pixel values to convert the density map to a probability distribution. Figure \ref{fig:maps} shows a sample fixation density along with its heatmap.
The original images of COCO-Search18 dataset are 1050 by 1680 pixels. For more efficient computation we resize all search images to 320 by 512 pixels.  

\begin{figure*}[h]
\centering
\includegraphics[width = \textwidth]{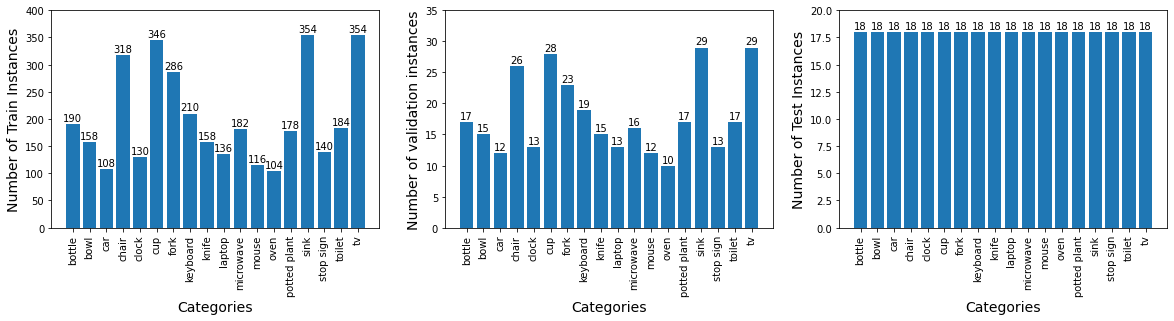}
\caption{Distribution of task-image instances across target categories for train, validation, and test splits.}
\label{fig:task-image}
\end{figure*}

We ignored the first fixation point in each trial, as all observers were fixating on a dot in the center of the screen at the onset of displaying each search image. We also removed the trials with incorrect responses from our data, i.e. when observers reported the wrong object as their detected target. We use the same train/valid split as was used in \cite{Goal-Directed-scan} in our modeling. After removing the incorrect trials, there remained 2150 training and 324 validation task-image pairs in COCOSearch18 dataset. Some of the search images are used more than once for different search targets. Hence we defined task-image pairs to distinguish between the search targets in similar search images. From 2150 training task-image pairs, we then separate 324 pairs for the test dataset, such that there are 18 task-images from 18 different categories. Hence, our train-validation-test split contains 1826-324-324 task-image pairs.
We further augment the training data by horizontally flipping the search images along with their ground-truth fixation density maps. The reason we chose horizontal flipping was inspired by \cite{8866748}, which suggests that among common transformations such as cropping, flipping (mirroring), rotating, changing brightness, and shearing, horizontal flip causes the least change in the fixation maps. After the augmentation, the train-validation-test split becomes 3652-324-324 task-image pairs. A histogram of the number of task-image instances for train, validation, and test splits over target categories can be seen in figure \ref{fig:task-image}.

\subsection{Training}

We initialize the VGG16 network with its pre-trained weights on ImageNet dataset for object recognition task, and then fine-tune (by adjusting all layers' weights) it on COCOSearch18 for our problem. It has been shown that high-level features extracted from object recognition networks can improve the performance of free-viewing saliency prediction. We therefore apply the same technique for visual search saliency. The weights of VGG networks used in both streams are shared. This weight sharing reduces the network's complexity and training time significantly, while also improving the accuracy. 

We train the network on COCOSearch18 dataset by minimizing KL-divergence loss in equation \ref{eq:kl_loss} between the predicted and ground-truth fixation density maps. We use the Adam optimizer with learning rate of $10^{-5}$, and train the network for 8 epochs, with batch size of 1. The input search images are all resized to 320 × 512, and the target images are resized to 64 × 64. The output fixation density map dimension is also 320 × 512 same as the input search image.  

\begin{equation}
\begin{aligned}
    D_{KL}(P||Q) = \sum_{i} Q_{i} \ln{(\epsilon + \frac{Q_i}{\epsilon + P_i})}
    \end{aligned}
    \label{eq:kl_loss}
\end{equation}

\subsection{Experiments and Analysis }

We evaluate our model using 8 saliency metrics: area under ROC curve (AUC Judd), AUC Borji, shuffled AUC (sAUC), normalized scanpath saliency (NSS), Kullback–Leibler divergence (KLD),  Pearson’s correlation coefficient (CC), similarity measure (SIM), and information gain (IG). Among these metrics, NSS, IG, AUC Judd, AUC Borji, and sAUC are considered location-based, while KLD and CC are distribution-based \cite{8315047}. For location-based metrics, groundtruth \textbf{binary} fixation maps are used, and for distribution-based metrics, groundtruth \textbf{blurred} fixation maps (convolved with Gaussian kernel densities) are used in evaluation. 
To compute KLD, the predicted and groudtruth blurred fixation maps are first converted to probability distributions, using a division by the sum of pixel values. Then formula \ref{eq:kl_loss} is used to compute the dissimilarity between the groundtruth and prediction. IG needs the predicted and baseline saliency maps to be normalized between 0 and 1 (using min-max normalization) and then divided by their sum to be converted to a probability distribution. Baseline saliency map contains all blurred fixations from all other images (overlayed and normalized). Then for all fixation locations in the groundtruth binary fixation map, we calculate the information gain between the predicted and baseline saliency map. To compute SIM we also normalize the maps between 0 and 1 (to avoid a erroneous performance boost when a model assigns a nonzero value to every pixel) and divide them by their sum in order to ensure that the maximum value of score will be 1. Then we compute histogram intersection between the groundtruth and prediction. In NSS computation, the predicted maps are normalized such that their mean is zero and their standard deviation is 1. Then at the fixation locations in the groundtruth binary fixation maps, we compute the mean value of the normalized predicted saliency map. Similarly, in computing CC we normalize the prediction and groundtruth blurred fixation maps to have zero mean and standard deviation of 1, then we compute the correlation (dependence) between the groundtruth and prediction using formula \ref{eq:cc}.

\begin{equation}
\begin{aligned}
    CC = \frac{\sum_{i}{(P_{i}Q_{i})}}{\sqrt{\sum_{j}{(P_{j}^2 + Q_{j}^2)}}}
    \end{aligned}
    \label{eq:cc}
\end{equation}

AUC (Judd), AUC Borji, and Shuffled AUC all first normalize the predicted maps between 0 and 1 (using min-max normalization). AUC Judd computes the area under the ROC curve created by sweeping through threshold values determined by the range of saliency map values at fixation locations. AUC Borji and sAUC compute the area under ROC curve created by sweeping through threshold values at fixed step size until the maximum saliency map value is reached. Another difference between these three variations of AUC comes from the definition of false positive rate. AUC (Judd) computes the ratio of the non-fixated image pixels that have values above threshold in the predicted saliency maps. AUC Borji computes the the same ratio but instead of computing at all the non-fixated image pixels, it generates N splits (by default 100) of random locations sampled uniformly from all image pixels (the number of random locations in each split is the same as the number of fixation locations), then computes the ratio for each split and averages the ratios over the N split. Shuffled AUC computes false positive ratio similar to AUC-Borji, but instead of randomly choosing from all image pixels, it uniformly samples from binary fixation maps of other images (we use only one other fixation map in our implementation) to create the splits and compute the average ratio. 

As in \cite{KRONER2020261}, we also evaluated our predicted fixation distributions without applying any metric-specific post-processing methods. In order to obtain an accurate and loss-less evaluation, we store the predicted and groundtruth fixation maps as numpy files instead of images. The results of the evaluation on unseen test set extracted from COCOSearch18 dataset is summarized in table \ref{table:results}. The overall performance of our 2-stream model is obtained on all target categories on a test set with 324 images, containing 18 sample images for each of the 18 target categories. We further compare the performance of our unified 2-stream model with single category-specific networks trained separately for each target category. Our current unified model has two streams that allow it to receive information about the target category, and output target-specific FDM predictions based on the target. An alternative approach is to train separate one-stream networks for each of the 18 target categories. As these network are inherently target-specific, they do not need a separate stream for a sample target. For a better comparison, we report the results of our two-stream network on test images of separate categories (18 test images for each category) along with the results of single category-specific trained models in table \ref{table:results_1}, \ref{table:results_2}, and \ref{table:results_3}. Although in some cases, one-stream models trained on one category achieve slightly better performance, a unified two-stream model that works for all categories demands less memory, is easier to train, and is more biologically plausible. Also, as the weights between the target and image streams are shared, a two stream model has the same level of complexity and number of parameters as the one-stream network. A slight increase in the training time of the 2-stream model is caused by a bigger training and validation set at each epoch, as we train on all images rather than only images of a single category. We also took the average performance of our category-specific 1-stream models over 18 target categories, and report the results in the second row of table \ref{table:results}. To clarify the impact of a separate target stream in our unified model, we report the results of a one-stream model trained and tested on all categories, in the third row of table \ref{table:results}. By comparing the first and third rows in table \ref{table:results}, we see a boost in performance of the model across all saliency metrics especially NSS, SIM, and IG.

\begin{table*}[htbp]
\captionsetup{justification=centering}
\caption{The average performance of the 2-stream and 1-stream category-specific models on COCOSearch18 test set for bottle, bowl, car, chair, clock, cup, fork, and keyboard categories are reported. The values are averaged over 5 independent runs. }

\makebox[\textwidth][c]{
    \begin{tabular}{ccccccccccc}
\toprule
\multirow{2}{*}{\textbf{Category}} & \multicolumn{2}{c}{\multirow{2}{*}{\textbf{Model}}}& \multicolumn{8}{c}{ \textbf{Saliency Metrics}}  \\

\cmidrule(lr){4-11}
&  &  & \textbf{AUC Judd} & \textbf{AUC Borji} & \textbf{sAUC} & \textbf{NSS} & \textbf{KLD} &\textbf{CC} & \textbf{SIM} & \textbf{IG}\\
\midrule

\multirow{2}{*}{\textbf{All}} & \multirow{2}{*}{\textbf{2-stream}} & \textbf{$\mu$} & \textbf{0.947} & \textbf{0.849} & \textbf{0.836} & \textbf{4.643} & \textbf{0.931} & \textbf{0.717} & \textbf{0.539} & \textbf{2.589}\\
&  & \textbf{$\sigma$} & 0.003 & 0.005 & 0.005 & 0.105 & 0.019 & 0.009 & 0.009 & 0.076\\
\midrule

\multirow{4}{*}{\textbf{Bottle}} & \multirow{2}{*}{\textbf{2-stream}} & \textbf{$\mu$} &\textbf{0.937}	&\textbf{0.837}	&\textbf{0.859}&\textbf{	4.238}&	0.965&	0.686&	0.512	&3.223\\
&  & \textbf{$\sigma$} &0.005&	0.020&	0.023&	0.364&	0.105&	0.042&	0.026&	0.602\\

\cmidrule(lr){2-11}
 & \multirow{2}{*}{\textbf{1-stream}} & \textbf{$\mu$} &0.935 & 0.837 & 0.828 & 3.952 & \textbf{0.941}& \textbf{0.706} & \textbf{0.523} & \textbf{3.485}\\
 &  & \textbf{$\sigma$} & 0.011	& 0.030	& 0.027	
 & 0.367 & 0.073 & 0.026 & 0.014 & 0.764\\
\midrule

\multirow{4}{*}{\textbf{Bowl}} & \multirow{2}{*}{\textbf{2-stream}} & \textbf{$\mu$}&0.925	&0.811	&\textbf{0.842}	&3.579	&1.144	&0.615	&0.453	&3.021\\
&  & \textbf{$\sigma$} &0.013	&0.031	&0.039	&0.468	&0.144&	0.070	&0.040	&0.765\\
\cmidrule(lr){2-11}
 & \multirow{2}{*}{\textbf{1-stream}} & \textbf{$\mu$} &\textbf{0.930} & \textbf{0.852} & 0.838 & \textbf{3.783} & \textbf{1.013}& \textbf{0.683} & \textbf{0.470} & \textbf{3.333}\\
 &  & \textbf{$\sigma$} & 0.004 & 0.013 & 0.034 & 0.271 & 0.076 & 0.020 & 0.025 & 0.744\\
 \midrule

\multirow{4}{*}{\textbf{Car}} & \multirow{2}{*}{\textbf{2-stream}} & \textbf{$\mu$} &0.916	&0.816	&0.849	&3.361	&1.337	&0.540	&0.415	&\textbf{3.953}\\
&  & \textbf{$\sigma$} & 0.017	&0.029	&0.030	&0.284	&0.157	&0.052	&0.036	&0.701\\
\cmidrule(lr){2-11}
 & \multirow{2}{*}{\textbf{1-stream}} & \textbf{$\mu$} & \textbf{0.941 }& \textbf{0.864} & \textbf{0.868} &\textbf{ 3.966} & \textbf{0.863}& \textbf{0.719} & \textbf{0.526} & 3.331\\
 &  & \textbf{$\sigma$} & 0.018 & 0.027 & 0.033 & 0.476 & 0.136 & 0.069 & 0.044 & 0.730\\
 \midrule

\multirow{4}{*}{\textbf{Chair}} & \multirow{2}{*}{\textbf{2-stream}} & \textbf{$\mu$} &0.907	&0.798	&0.837	&2.974	&1.225	&0.557	&0.444	&2.866\\
&  & \textbf{$\sigma$} &0.018	&0.030	&0.025	&0.495	&0.134	&0.065	&0.020	&0.662\\
\cmidrule(lr){2-11}
 & \multirow{2}{*}{\textbf{1-stream}} & \textbf{$\mu$} &\textbf{0.941} & \textbf{0.864} & \textbf{0.868} & \textbf{3.966} & \textbf{0.863}& \textbf{0.719} & \textbf{0.526} & \textbf{3.331}\\
 &  & \textbf{$\sigma$} & 0.016 & 0.031 & 0.052 & 0.388 & 0.130 & 0.056  & 0.027 & 0.175\\
 \midrule

\multirow{4}{*}{\textbf{Clock}} & \multirow{2}{*}{\textbf{2-stream}} & \textbf{$\mu$} &0.965	&0.881&	0.892&	7.252&	0.824&	0.821&	\textbf{0.609}&	4.838\\
&  & \textbf{$\sigma$} &0.005&	0.016&	0.018&	0.512&	0.065&	0.033&	0.049&	0.520\\

\cmidrule(lr){2-11}
 & \multirow{2}{*}{\textbf{1-stream}} & \textbf{$\mu$} &\textbf{0.972} & \textbf{0.899} & \textbf{0.899} & \textbf{7.450} & \textbf{0.700}& \textbf{0.863} & 0.593 & \textbf{4.871}\\
 &  & \textbf{$\sigma$} & 0.003&0.006&	0.011&	0.414&	0.039&	0.026&	0.028&	0.953\\
 \midrule
 
\multirow{4}{*}{\textbf{Cup}} & \multirow{2}{*}{\textbf{2-stream}} & \textbf{$\mu$} &0.937	&0.835	&\textbf{0.858}	&4.106	&1.025&	0.675&	0.515&\textbf{	3.331}\\
&  & \textbf{$\sigma$} &0.012&	0.024&	0.027&	0.351&	0.166&	0.057&	0.033&	0.915\\
\cmidrule(lr){2-11}
 & \multirow{2}{*}{\textbf{1-stream}} & \textbf{$\mu$} &\textbf{0.950} & \textbf{0.867} & 0.840 &\textbf{ 4.390} & \textbf{	0.833}&\textbf{ 0.764} &\textbf{ 0.541} & 3.233\\
 &  & \textbf{$\sigma$} &0.009	&0.020	&0.042	&0.322	&0.119	&0.045	&0.032	&0.235\\
 \midrule
 
\multirow{4}{*}{\textbf{Fork}} & \multirow{2}{*}{\textbf{2-stream}} & \textbf{$\mu$ }& 0.916	&0.794	&0.814	&4.079	&1.188&	0.666	&\textbf{0.497}&	\textbf{3.333}
\\
&  & \textbf{$\sigma$} &0.024&	0.034&	0.032&	0.706&	0.297&	0.107&	0.068&	0.202
\\
\cmidrule(lr){2-11}
 & \multirow{2}{*}{\textbf{1-stream}} & \textbf{$\mu$} & \textbf{0.929} &\textbf{0.835} &\textbf{0.824} &\textbf{4.518}	&\textbf{1.014}	&\textbf{0.714}	&0.489	&2.975\\
 &  & \textbf{$\sigma$} &0.014	&0.020	&0.044	&0.405	&0.110	&0.057	&0.026	&0.975\\
 \midrule
 
\multirow{4}{*}{\textbf{Keyboard}} & \multirow{2}{*}{\textbf{2-stream}} & \textbf{$\mu$} &0.955&	\textbf{0.874}&	\textbf{0.891}&	4.239&	\textbf{0.950}&	0.699&\textbf{	0.531}&	2.344\\
&  & \textbf{$\sigma$}&0.011&	0.035&	0.040&	0.374&	0.128&	0.043&	0.031&	0.790\\
\cmidrule(lr){2-11}
 & \multirow{2}{*}{\textbf{1-stream}} & \textbf{$\mu$} & \textbf{0.958}& 0.867 &0.736&	\textbf{4.486}&	0.979&\textbf{0.700}&	0.507	&\textbf{2.450}\\
 &  & \textbf{$\sigma$} &0.011	&0.029	&0.092	&0.427	&0.133	&0.054	&0.033	&0.604\\
 \midrule
\end{tabular}%
}

\label{table:results_1}

\end{table*}%

\begin{table*}[htbp]
\captionsetup{justification=centering}
\caption{The average performance of the 2-stream and 1-stream category-specific models on COCOSearch18 test set for knife, laptop, microwave, mouse, oven, potted plant, sink, and stop sign categories are reported. The values are averaged over 5 independent runs.}

\makebox[\textwidth][c]{
    \begin{tabular}{ccccccccccc}
\toprule
\multirow{2}{*}{\textbf{Category}} & \multicolumn{2}{c}{\multirow{2}{*}{\textbf{Model}}}& \multicolumn{8}{c}{ \textbf{Saliency Metrics}}  \\

\cmidrule(lr){4-11}
&  &  & \textbf{AUC Judd} & \textbf{AUC Borji} & \textbf{sAUC} & \textbf{NSS} & \textbf{KLD} &\textbf{CC} & \textbf{SIM} & \textbf{IG}\\
\midrule

\multirow{4}{*}{\textbf{Knife}} & \multirow{2}{*}{\textbf{2-stream}} & \textbf{$\mu$} &0.909	&0.785&	\textbf{0.818}	&2.969	&1.462	&0.513	&\textbf{0.411}	&\textbf{2.728}\\
&  & \textbf{$\sigma$} &0.018&	0.037&	0.038&	0.530&	0.195	&0.078&	0.053	&0.669\\
\cmidrule(lr){2-11}
 & \multirow{2}{*}{\textbf{1-stream}} & \textbf{$\mu$} & \textbf{0.910}&\textbf{0.821}&	0.797&	\textbf{3.029}&	\textbf{1.399}&	\textbf{0.531}&	0.387&	2.525\\
 &  & \textbf{$\sigma$} & 0.018	&0.042	&0.058	&0.511	&0.159	&0.073	&0.042	&0.865\\
\midrule


\multirow{4}{*}{\textbf{Laptop}} & \multirow{2}{*}{\textbf{2-stream}} & \textbf{$\mu$} &\textbf{0.949}	&\textbf{0.851}	&\textbf{0.871}	&\textbf{4.176}	&\textbf{0.984}	&\textbf{0.690}	&\textbf{0.519}	&\textbf{2.817}\\
&  & \textbf{$\sigma$} &0.011	&0.022	&0.023	&0.362	&0.131	&0.043	&0.023	&0.444\\
\cmidrule(lr){2-11}
 & \multirow{2}{*}{\textbf{1-stream}} & \textbf{$\mu$} &0.936&	0.841&	0.812&	3.992	&1.113&	0.632&	0.454&	2.317\\
 &  & \textbf{$\sigma$} & 0.012	&0.023	&0.035	&0.468  & 0.0465	&0.028	&0.025	&0.461\\
 \midrule

\multirow{4}{*}{\textbf{Microwave}} & \multirow{2}{*}{\textbf{2-stream}} & \textbf{$\mu$} &0.958	&0.857&	\textbf{0.871}&	4.743&	\textbf{0.757}&	0.759&	\textbf{0.579}&	2.964
\\
&  & \textbf{$\sigma$} &0.005&	0.019&	0.024&	0.255&	0.064&	0.025&	0.025&	0.650\\
\cmidrule(lr){2-11}
 & \multirow{2}{*}{\textbf{1-stream}} & \textbf{$\mu$} &\textbf{0.962}	&\textbf{0.872}	&0.862	&\textbf{4.782}	&0.696	&\textbf{0.789}	&\textbf{0.579}	&\textbf{3.205}\\
 &  & \textbf{$\sigma$} & 0.003	&0.014	&0.037	&0.213	&0.025	&0.017	&0.013	&0.269\\
 \midrule


\multirow{4}{*}{\textbf{Mouse}} & \multirow{2}{*}{\textbf{2-stream}} & \textbf{$\mu$} &\textbf{0.944}&	0.831&	\textbf{0.851}&	4.595&	\textbf{1.040}&	0.682&	\textbf{0.504}&	\textbf{3.292}\\
&  & \textbf{$\sigma$} &0.011&	0.021&	0.017&	0.734&	0.203&	0.077&	0.070&	0.432\\
\cmidrule(lr){2-11}
 & \multirow{2}{*}{\textbf{1-stream}} & \textbf{$\mu$} & 0.943&\textbf{0.840}	&0.839	&\textbf{4.884}  &1.051 &\textbf{0.738}	&0.499	&3.091\\
 &  & \textbf{$\sigma$} & 0.010	&0.023	&0.021	&0.268	&0.055	&0.041	&0.018	&0.692\\
 \midrule
 

\multirow{4}{*}{\textbf{Oven}} & \multirow{2}{*}{\textbf{2-stream}} & \textbf{$\mu$} &\textbf{0.959}&	0.864&	\textbf{0.883}&	\textbf{4.327}&	0.888&	0.713&	\textbf{0.545}&\textbf{	2.588}\\
&  & \textbf{$\sigma$} &0.004&	0.017&	0.021&	0.278&	0.144&	0.037&	0.028&	0.533\\
\cmidrule(lr){2-11}
 & \multirow{2}{*}{\textbf{1-stream}} & \textbf{$\mu$} &0.957 &\textbf{0.886}	&0.853	&4.304	&\textbf{0.814}	&\textbf{0.745}	&0.537	&2.324\\
 &  & \textbf{$\sigma$} & 0.006&	0.023&	0.065&	0.319&	0.110&	0.050&	0.037&	0.582\\
 \midrule

\multirow{4}{*}{\textbf{Potted plant}} & \multirow{2}{*}{\textbf{2-stream}} & \textbf{$\mu$} & \textbf{0.944}	&\textbf{0.856}&	\textbf{0.880}&	\textbf{4.487}&	\textbf{0.849}&	\textbf{0.742}&	\textbf{0.545}&\textbf{	3.950}\\
&  & \textbf{$\sigma$}&0.009&	0.021&	0.024&	0.543&	0.091&	0.060&	0.046&	0.689\\
\cmidrule(lr){2-11}
 & \multirow{2}{*}{\textbf{1-stream}} & \textbf{$\mu$} &0.943 &\textbf{0.856} &0.853	&4.005&0.908	&0.708&0.504	&3.303\\
 &  & \textbf{$\sigma$} &0.013	&0.031	&0.038	&0.216	&0.077	&0.032	&0.016	&0.507\\
 \midrule

\multirow{4}{*}{\textbf{Sink}} & \multirow{2}{*}{\textbf{2-stream}} & \textbf{$\mu$} &0.957	&0.872	&\textbf{0.891}	&\textbf{4.683}&	0.749&	\textbf{0.773}&	\textbf{0.581}&	\textbf{3.686}\\
&  & \textbf{$\sigma$} &0.016&	0.034	&0.031&	0.477&	0.205&	0.066&	0.045&	0.408\\
\cmidrule(lr){2-11}
 & \multirow{2}{*}{\textbf{1-stream}} & \textbf{$\mu$} &\textbf{0.959}	&\textbf{0.884}&	0.874&	4.454&	\textbf{0.745}	&0.754&	0.562&3.324\\
 &  & \textbf{$\sigma$} &0.008	&0.017	&0.023	&0.362	&0.104	&0.039	&0.034	&0.916\\
 \midrule

\multirow{4}{*}{\textbf{Stop sign}} & \multirow{2}{*}{\textbf{2-stream}} & \textbf{$\mu$} &\textbf{0.968}&	0.891&	\textbf{0.904}&\textbf{	6.694}&\textbf{	0.777}&	\textbf{0.821}&	\textbf{0.620}&	\textbf{5.215}
\\
&  & \textbf{$\sigma$} &0.006&	0.020&	0.023&	0.443&	0.119&	0.036&	0.029&	0.292\\
\cmidrule(lr){2-11}
 & \multirow{2}{*}{\textbf{1-stream}} & \textbf{$\mu$} & 0.965	&\textbf{0.892}	&0.870&6.176&	0.788  &0.814&	0.591&4.791\\
 &  & \textbf{$\sigma$} &0.009	&0.018	&0.035&	0.413&	0.116&	0.025&	0.024&	0.992\\
 \midrule

\end{tabular}%
}

\label{table:results_2}

\end{table*}%

\begin{table*}[htbp]
\captionsetup{justification=centering}
\caption{The average performance of the 2-stream and 1-stream category-specific models on COCOSearch18 test set for toilet and TV categories are reported. The values are averaged over 5 independent runs.}

\makebox[\textwidth][c]{
    \begin{tabular}{ccccccccccc}
\toprule
\multirow{2}{*}{\textbf{Category}} & \multicolumn{2}{c}{\multirow{2}{*}{\textbf{Model}}}& \multicolumn{8}{c}{ \textbf{Saliency Metrics}}  \\

\cmidrule(lr){4-11}
&  &  & \textbf{AUC Judd} & \textbf{AUC Borji} & \textbf{sAUC} & \textbf{NSS} & \textbf{KLD} &\textbf{CC} & \textbf{SIM} & \textbf{IG}\\
\midrule

\multirow{4}{*}{\textbf{Toilet}} & \multirow{2}{*}{\textbf{2-stream}} & \textbf{$\mu$} &\textbf{0.967}&	0.872&	0.888&	\textbf{5.014}&	\textbf{0.774}&	\textbf{0.776}&	\textbf{0.573}&	\textbf{3.649}\\
&  & \textbf{$\sigma$} &0.008&	0.023&	0.025&	0.488&	0.163&	0.058&	0.043&	0.487\\
\cmidrule(lr){2-11}
 & \multirow{2}{*}{\textbf{1-stream}} & \textbf{$\mu$} &0.965	&\textbf{0.891}	&\textbf{0.893 }&4.724	&0.851 &0.761	&0.534 &	3.583\\
 &  & \textbf{$\sigma$} &0.003	&0.008	&0.006	&0.159	&0.023	&0.014	&0.020	&1.405\\
 \midrule

\multirow{4}{*}{\textbf{TV}} & \multirow{2}{*}{\textbf{2-stream}} & \textbf{$\mu$} &0.965&	0.871&	\textbf{0.878}&	5.704&	0.735&	0.818&	\textbf{0.623}&\textbf{	4.040}\\
&  & \textbf{$\sigma$} &0.007&	0.028&	0.021&	0.252&	0.138&	0.026&	0.018&	0.645\\
\cmidrule(lr){2-11}
 & \multirow{2}{*}{\textbf{1-stream}} & \textbf{$\mu$} & \textbf{0.966}	&\textbf{0.877}	&0.873	&\textbf{5.873}	&\textbf{0.675}&\textbf{0.827}	&0.611 & 3.274\\
 &  & \textbf{$\sigma$} &0.013&	0.038	&0.040&	0.698&	0.157&	0.048&	0.052&	0.254\\
 \midrule

\end{tabular}%
}
\label{table:results_3}
\end{table*}%

\begin{figure*}[p]
\centering
\includegraphics[width=14cm,height=22cm]{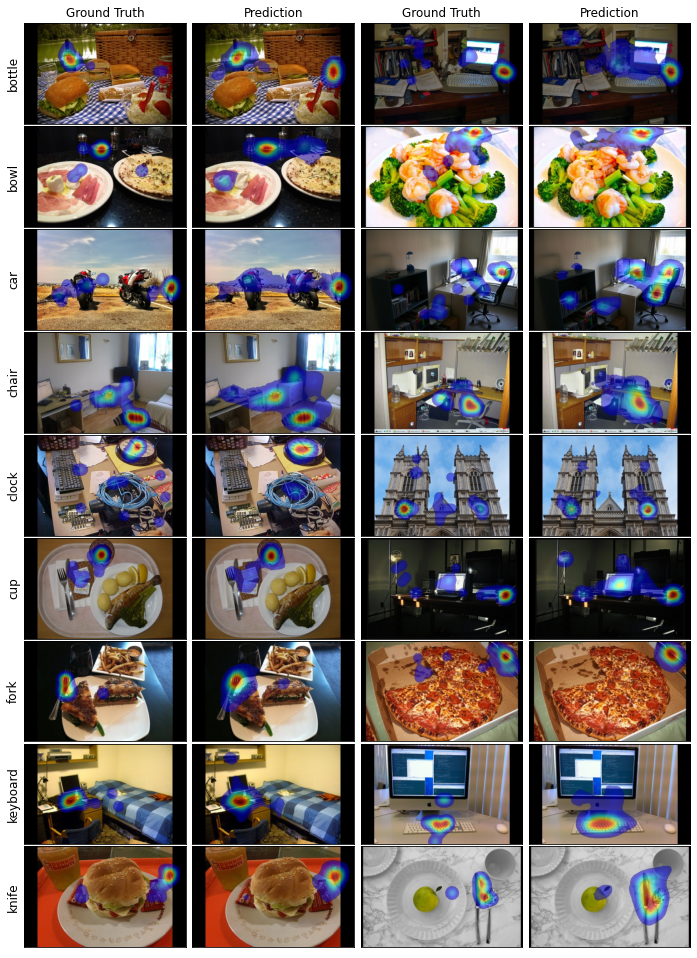}
\caption{A visualization of network's predictions and ground-truth fixation density maps on unseen test data for 9 categories. The target category is specified on the left of each row.}
\label{fig:results_1}
\end{figure*}

\begin{figure*}[p]
\centering
\includegraphics[width=14cm,height=22cm]{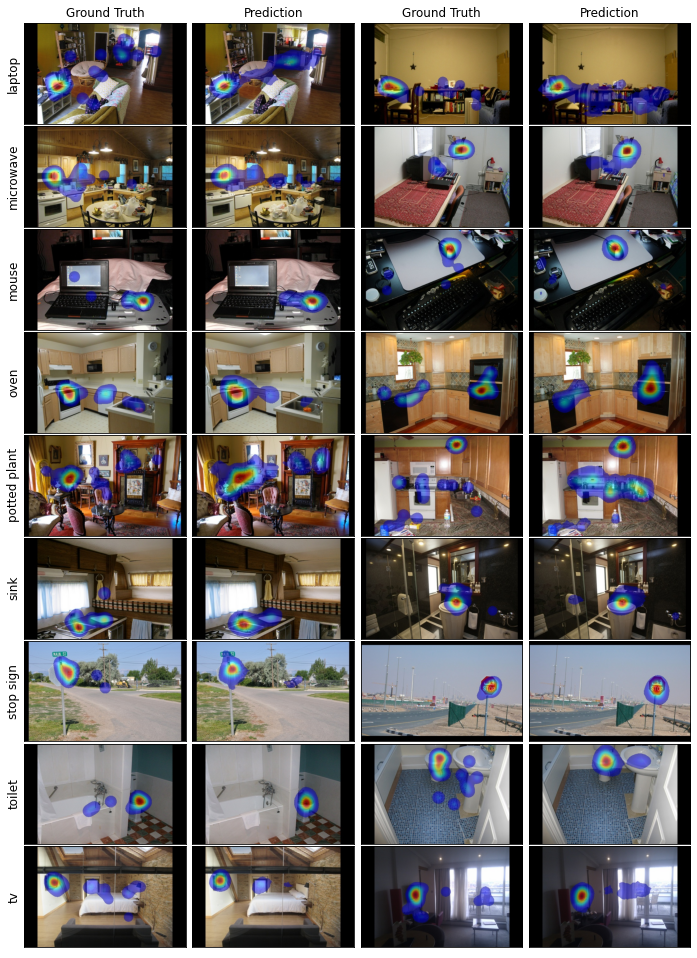}
\caption{A visualization of network's predictions and ground-truth fixation density maps on unseen test data for the remaining 9 categories.}
\label{fig:results_2}
\end{figure*}

To gain insight into the qualitative performance of our model, we visualized the network's predicted density maps and compared them to the ground-truth density maps for the test dataset. Some of these visualizations are shown in figures \ref{fig:results_1} and \ref{fig:results_2}. The qualitative results show that the network accomplishes to detect the targets in most of the images. It also learns where to look for each target category based on its experience in the training data. For instance, when looking for oven or toilet, the locations closer to the ground are considered salient; however, for bottle, bowl, laptop, fork, knife, mouse, and cup, the top of surfaces such as tables are mostly salient. While searching for TV and clock the higher items in images are often considered salient. Moreover, the items sharing some features with the objects of target category are considered salient both by the network and human observers. As an example, when observers are looking for a bottle, objects such as glasses and cups that are also relatively cylindrical, are fixated. In general, the network's predictions and ground-truth FDMs have great similarities with matching distraction patterns.

\begin{table*}[htbp]
\captionsetup{justification=centering}
\caption{The average performance of the model on randomly selected test set. The values are averaged over 5 independent runs.}

\makebox[\textwidth][c]{
    \begin{tabular}{ccccccccccc}
\toprule
\multirow{2}{*}{\textbf{Category}} & \multicolumn{2}{c}{\multirow{2}{*}{\textbf{Model}}}& \multicolumn{8}{c}{ \textbf{Saliency Metrics}}  \\

\cmidrule(lr){4-11}
&  &  & \textbf{AUC Judd} & \textbf{AUC Borji} & \textbf{sAUC} & \textbf{NSS} & \textbf{KLD} &\textbf{CC} & \textbf{SIM} & \textbf{IG}\\
\midrule

\multirow{2}{*}{\textbf{All}} & \multirow{2}{*}{\textbf{2-stream}} & \textbf{$\mu$} &\textbf{0.947}	&0.849	&0.836	&\textbf{4.643}	&0.931	&0.717	&\textbf{0.539}	&2.589\\
&  & \textbf{$\sigma$} & 0.003	&0.005	&0.005	&0.105	&0.020	&0.009	&0.009	&0.076\\
 \midrule

\multirow{2}{*}{\textbf{Single}} & \multirow{2}{*}{\textbf{1-stream}} & \textbf{$\mu$} & 0.946	&\textbf{0.862}&	\textbf{0.843}&	4.552&	\textbf{0.916}&	\textbf{0.725}&	0.520&	\textbf{3.247}
\\
&  & \textbf{$\sigma$} & 0.018	&0.023	&0.0387	&1.063	&0.184	&0.080	&0.058	&0.696\\
\midrule

\multirow{2}{*}{\textbf{All}} & \multirow{2}{*}{\textbf{1-stream}} & \textbf{$\mu$} &0.923	&0.820&	0.804&	3.309	&1.361	&0.515	&0.392	&1.901\\
&  & \textbf{$\sigma$} &0.002&	0.010&	0.007&	0.071&	0.042	&0.014	&0.010	&0.042\\
\midrule

\end{tabular}%
}
\label{table:results}
\end{table*}%

\begin{table*}[htbp]
\captionsetup{justification=centering}
\caption{A comparison between the performance of the model on unseen test data of COCOSearch-18 dataset, with and without data augmentation. The values are averaged over 4 independent runs.}

\makebox[\textwidth][c]{
    \begin{tabular}{cccccccccc}
\toprule
 \multicolumn{2}{c}{\multirow{2}{*}{\textbf{Model}}}& \multicolumn{8}{c}{ \textbf{Saliency Metrics}}  \\

\cmidrule(lr){3-10}
&  & \textbf{AUC Judd} & \textbf{AUC Borji} & \textbf{sAUC} & \textbf{NSS} & \textbf{KLD} &\textbf{CC} & \textbf{SIM} & \textbf{IG}\\
\midrule

\multirow{2}{*}{\textbf{With Augmentation}} & \textbf{$\mu$} &\textbf{0.947}	&0.849	&0.836	&\textbf{4.643}	&\textbf{0.931}	&\textbf{0.717}&	\textbf{0.539}&	\textbf{2.589}\\
& \textbf{$\sigma$} & 0.003	&0.005&	0.005&	0.105&	0.020&	0.009&	0.009&	0.076\\
\midrule

\multirow{2}{*}{\textbf{Without Augmentation}} & \textbf{$\mu$} &0.943	&\textbf{0.854}	&\textbf{0.845}	&4.499	&0.977	&0.695	&0.513	&2.503\\
& \textbf{$\sigma$} & 0.003	&0.013	&0.015	&0.096	&0.030	&0.009	&0.009	&0.039\\
 \midrule

\end{tabular}%
}
\label{table:augment}
\end{table*}%

\begin{table*}[htbp]
\captionsetup{justification=centering}
\caption{A comparison between the performance of the model on unseen test data of COCOSearch-18 dataset, with and without ASPP architecture. The values are averaged over 4 independent runs.}

\makebox[\textwidth][c]{
    \begin{tabular}{cccccccccc}
\toprule
 \multicolumn{2}{c}{\multirow{2}{*}{\textbf{Model}}}& \multicolumn{8}{c}{ \textbf{Saliency Metrics}}  \\

\cmidrule(lr){3-10}
&  & \textbf{AUC Judd} & \textbf{AUC Borji} & \textbf{sAUC} & \textbf{NSS} & \textbf{KLD} &\textbf{CC} & \textbf{SIM} & \textbf{IG}\\
\midrule

\multirow{2}{*}{\textbf{With ASPP}} & \textbf{$\mu$} &\textbf{0.947}	&\textbf{0.849}	&0.836	&\textbf{4.643}	&\textbf{0.931}	&\textbf{0.717}&	\textbf{0.539}&	\textbf{2.589}\\ 
& \textbf{$\sigma$} & 0.003	&0.005&	0.005&	0.105&	0.020&	0.009&	0.009&	0.076\\ 
\midrule 
\multirow{2}{*}{\textbf{Without ASPP}} & \textbf{$\mu$} &0.942	&0.846	&\textbf{0.838}   &4.579	&0.975	&0.712	&0.527	&2.463\\
& \textbf{$\sigma$} & 0.002	&0.006	&0.008	&0.083	&0.041	&0.014	&0.011	&0.070\\ 
 \midrule 

\end{tabular}%
}
\label{table:aspp}
\end{table*}%

\begin{figure*}
    \centering
    \subfigure[bottle]{
    \centering
        \includegraphics[width=0.3\textwidth]{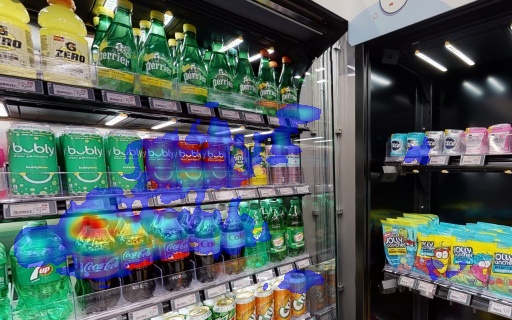}
    }
    \subfigure[bottle]{
    \centering
          \includegraphics[width=0.3\textwidth]{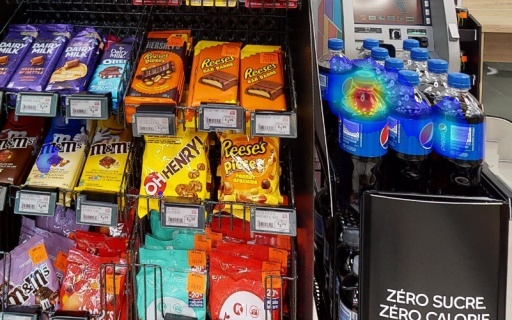}
    }
    \subfigure[bottle]{
    \centering
          \includegraphics[width=0.3\textwidth]{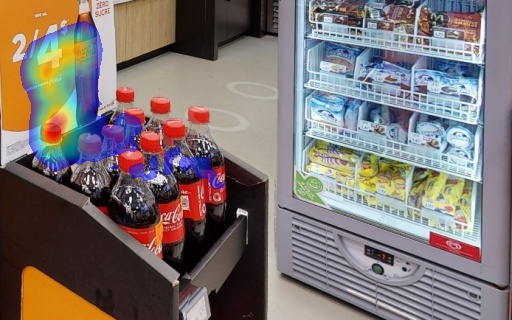}
   }
    \subfigure[bottle]{
    \centering
          \includegraphics[width=0.3\textwidth]{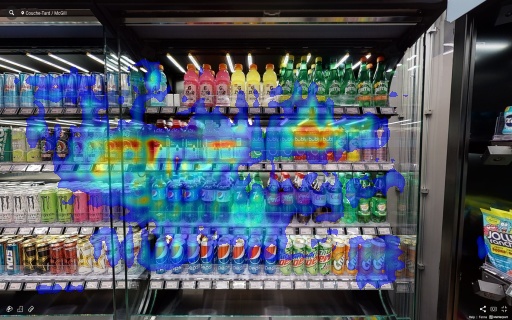}
   }
    \subfigure[bowl]{
    \centering
          \includegraphics[width=0.3\textwidth]{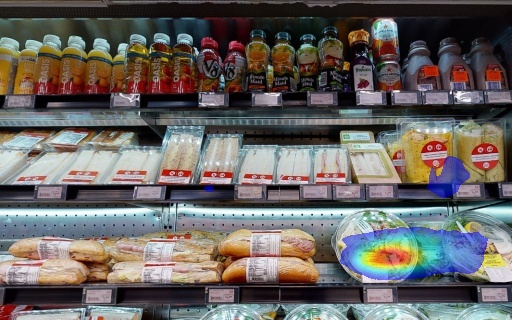}
    }
    \subfigure[cup]{
    \centering
          \includegraphics[width=0.3\textwidth]{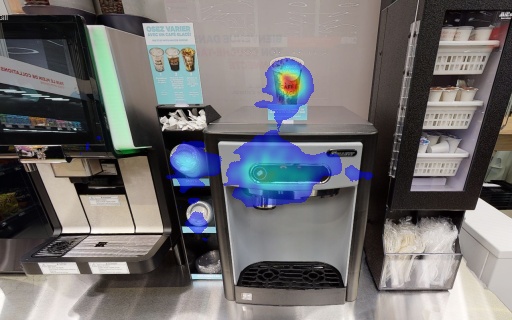}
    }
    \subfigure[microwave]{
    \centering
          \includegraphics[width=0.3\textwidth]{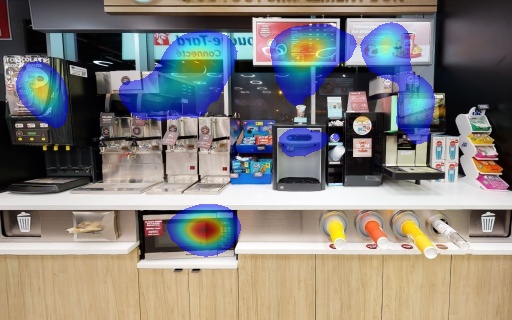}
   }
\caption{Network predictions on Couche-Tard store images. The target object category of each image is listed below it. As before, for each target search category the image fed into the target stream of network is randomly chosen from images shown in \ref{fig:targets}.}
\label{fig:store}
\end{figure*}

We also tested our network on the images captured from Couche-Tard laboratory store at McGill campus. Some of these images are shown in figure \ref{fig:store}. We do not have the ground truth eye fixation data on these images to compare with our predictions; however, it can be seen that the network generates qualitatively reasonable results, especially for images which possess the same level of clutter as COCOSearch18 dataset. In figures \ref{fig:store}-a and \ref{fig:store}-d the level of clutter is very high and all of the objects also belong to the `bottle' target category. This causes the network to predict a large salient region for these images. Also, our network is mostly trained on images that only contain one instance of the target category, thus it often fails to detect all targets as salient when more than one target object is present in the image. We can see this behavior in figures \ref{fig:store}-c and \ref{fig:store}-b, where only one instance of target is selected as salient. In figure \ref{fig:store}-c, the network makes a reasonable prediction that the observer will be distracted by the discount offer on coca-colas. In figure \ref{fig:store}-f, the network predicts that when the observers are searching for the cup they most likely get distracted by the circular plastic caps on the side of the espresso machine. In figure \ref{fig:store}-g, the network predicts most of the objects which have the same shape as a microwave, have steel cover, and are located higher in the image, as distracting. 

As an ablation study, we investigate the effect of data augmentation, presence of ASPP, presence of decoder, initializing VGG with random weights, and not sharing weights between the target and search image streams.

We realized that with data augmentation, using horizontal flipping, the performance of the model increased on most saliency metrics, as can be seen in table \ref{table:augment}. The average AUC-Borji, and sAUC are slightly lower for augmented model, but considering the standard deviation of the results they seem to be around the same range. Adding more augmentations such as vertical flipping did not lead to performance increase. The reason might be due to the effect of vertical flipping transformation on the fixation pattern. The fixation map of a vertically flipped image might no longer be in accordance with the vertically flipped fixation map of the original image. This would create erroneous data, which causes the model's performance to drop.

To explore the effect of ASPP architecture, we removed it and directly fed the concatenated VGG16 features to a 1 × 1 convolution with 256 filters. Then we passed the output through the decoder. As can be seen in table \ref{table:aspp}, the results validate the effectiveness of ASPP architecture in feature extraction.

As a second architecture change, we removed the decoder from the network, and up-sampled the output of the convolved streams by 8 in one shot. Then we normalized the output to create a probabilistic density map. Removing the decoder leads to checkerboard artifacts in the predicted density maps, and the maps seem to be more sparse. 

We realized that initializing the weights of the feature extractor backbone network, i.e. VGG16, randomly instead of using pretrained weights on ImageNet, dramatically decreases the performance. 

We performed vast experiments on the size of the input sample target image, and chose 64x64 dimension as the optimal resolution.

We further investigate whether pre-training the one-stream category-specific models on SALICON free-viewing dataset can enhance their performance on COCOSearch18. Unlike the free-viewing scenario where pre-training on SALICON dataset leads to performance boost, in case of visual search saliency, pretraining on SALICON degrades the performance. The main reason comes from the significant difference between a visual search FDM and free-viewing FDM. In the free-viewing condition, the gaze locations are more spread out over the image, while in visual search gaze is more focused on the target object. Also SALICON has 10000 training samples, almost three times larger than our augmented dataset. Hence, retraining the network on COCOSearch18 after it is pre-trained on SALICON does not change its behavior significantly, and the predicted FDMs are too scattered.

\begin{figure*}[p]
\centering
\includegraphics[width=14cm,height=22cm]{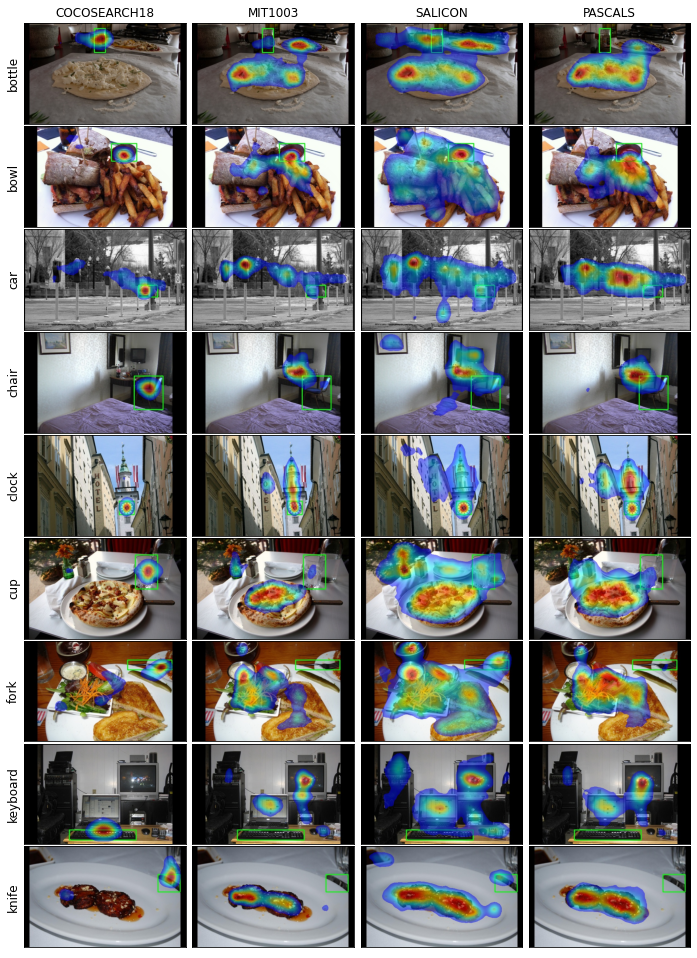}
\caption{Comparison of our visual search model predictions (trained on cocosearch18), and vanilla MSI-Net trained on three free-viewing datasets namely MIT-1003, salicon, and pascals, on the first 9 target categories.}
\label{fig:freeviewing}
\end{figure*}

\begin{figure*}[p]
\centering
\includegraphics[width=14cm,height=22cm]{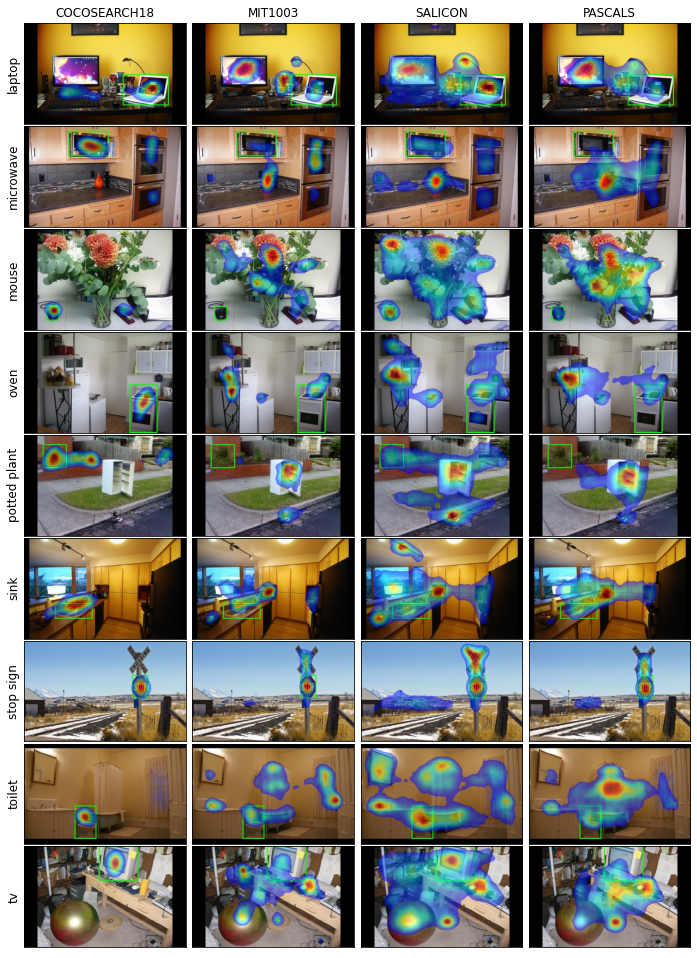}
\caption{Comparison of our visual search model predictions (trained on cocosearch18), and vanilla MSI-Net trained on three free-viewing datasets namely MIT-1003, salicon, and pascals, on the remaining 9 target categories.}
\label{fig:freeviewing_2}
\end{figure*}

To understand how our network predictions for visual search differ from a one-stream MSI-Net that is trained for free-viewing task, we tested free-viewing MSI-Net, trained separately on Salicon, MIT1003, and Pascals datasets, on COCOSearch18 dataset. The results are shown in figures \ref{fig:freeviewing} and \ref{fig:freeviewing_2}. In general, a vanilla MSI-Net trained on free-viewing datasets predicts a larger part of images as salient compared to our double-stream visual search model. Sometimes, the distractor objects that our network predicts as salient is also considered salient under free-viewing condition; however, this is not always the case. Also, in freeviewing saliency, the target objects are not always parts of the salient regions. 

Same as MSI-Net, our model is relatively light-weight. As the weights are shared between the two streams of our network, our model has 24,934,209 parameters similar to the one-stream MSI-Net. MSI-Net is among the lightest free-viewing saliency models, and here we exploit this feature by extending MSI-Net to a visual search saliency predictor.


\subsection{Discussion}

To the best of our knowledge, the model we proposed here is the first attempt to predict distraction through estimating fixation density maps for visual search using COCOSearch18 dataset. Although the creators of COCOSearch18 have partially worked on fixation density maps in \cite{Goal-Directed-scan} and \cite{cocos}, their main concern is predicting sequential scan-paths.

One failure of our proposed model is its confusion in identifying similar target categories such as fork and knife. In most images where a fork and knife are both present, the network mistakes one target with the other. Also our network sometimes generates high fixation probabilities for non-target objects that share close similarity (such as similar color or shape) with the target. Hence the FDM peaks will be mistakenly placed on these non-target objects. This behavior has been also seen in the ground-truth data of human observers. Many of the fixated non-target objects by human observers have resemblances to the target. In most cases, humans are able to find the target after fixating on the similar non-target objects; while the network sometimes fails to do that.

One direction for future research could be on the choice of the sample target images. Convolutional neural networks are not inherently rotation-invariant. Thus, the difference between the orientation of the sample target image and the target object in the search image could affect the modeling performance. One possible solution might be to use a rotation-invariant convolutional neural network such as RIFD-CNN \cite{7780684} in the target stream to extract rotation invariant features from the sample target images.

An application of our model could be in object detection. Predicting salient regions of a scene while searching for an object category can be used as an object proposal strategy in object detection algorithms. Salient regions highlight potential locations that might contain the target category in an image.


\section{Method 2: Predicting Segmentation of Targets and Distractors}

\label{sec:method_2}

In our second method, we aim to segment target and distractors during visual search. Our method involves training a separate Mask-RCNN instance segmentation network for each target category. In our modeling, we treat all pixels belonging to the same object as being equally distracting. This method is in accordance with object-based visual attention theories.

\subsection{Model Architecture}

We use Mask-RCNN \cite{he2018mask} instance segmentation network in our modeling. Mask-RCNN belongs to the RCNN (Regions with CNN features) group of networks. The first RCNNs, namely RCNN \cite{RCNN_}, fast-RCNN \cite{fastRCNN}, and faster RCNN \cite{FasterR-CNN}, were solely for object detection purposes. Mask-RCNN on the other hand provides pixel-level segmentation, by adding a new branch to faster-RCNN along with some modifications. We first describe faster-RCNN architecture and then extend that to Mask-RCNN. 

As the first step, faster-RCNN extracts multi-scale image features using FPN (Feature Pyramid Networks) \cite{lin2017feature} architecture. FPN consists of a bottom-up pathway, and a top-down architecture with lateral connections. The bottom-up pathway consists of popular CNN architectures such as ResNet or VGG, which extracts features from input images. In our modeling, we use ResNet-101 as our feature extraction backbone. Top-down pathway generates feature pyramid map, which has the same size as the bottom-up pathway. To create the pyramid, as we go down the top-down path, we upsample the previous layer by 2 using nearest neighbors upsampling. We apply a 1 × 1 convolution to the corresponding feature maps in the bottom-up pathway, and add it to the upsampled previous layer element-wise. A 3 × 3 convolution is then applied to all merged layers, which reduces the aliasing effect when merged with the upsampled layer. FPN outperforms single CNNs due to creating high-level semantic feature maps at various scales.

As the next step, the network proposes several boxes in the image and checks if any of them corresponds to an object. For this purpose, faster-RCNN has a fully convolutional network, known as region proposal network (RPN), on top of the extracted CNN features. This network passes a sliding window over the CNN feature map; at each window outputting k potential bounding boxes and the scores of how good each bounding box is. The k proposed bounding boxes are compared to k reference boxes, which are called anchors. Each anchor is centered at the sliding window in question, and has a scale and aspect ratio. At each sliding position, there are typically 9 anchors with 3 different scales and 3 aspect ratios. To generate features for each of these proposed bounding boxes, the pre-computed CNN features of the whole image are reused through a method called ROI pooling. Then for each bounding box, we pass its features into faster RCNN to generate a classification and run a linear regression to fit a tighter bounding box to the object.

To generate pixel-wise segmentation along with object classification and bounding box detection, authors introduced Mask-RCNN. In mask-RCNN, a fully convolutional neural network is added to the top of CNN features to generate a mask segmentation output. This is in parallel to the classification and bounding box regressor network of the faster RCNN. Also, authors realized that the regions of the feature map selected by ROI pooling were slightly misaligned from the regions of the original image. Since the image segmentation required pixel-level specificity unlike bounding boxes these misalignments led to inaccuracies. The authors solved this problem by introducing a method called ROI Align, which forces the cell boundaries of the target feature map to realign with the boundary of the input feature maps.

\subsection{Data Preprocessing}

COCOSearch-18 dataset does not include the segmentation of objects. Although all images in COCOSearch-18 are from the COCO dataset, which provides partial segmentation of objects, most of these segmentations contain neither the target nor the fixated distractors. Hence, we use COCO Annotator \cite{cocoannotator}, an online annotation platform, to manually segment the fixated objects. Due to time constraint, we have annotated 3 search categories: bottle, bowl, and car, out of 18 available categories. The output of COCO Annotator is a json file similar to the format of COCO dataset. For more information on the content of the json files please refer to \cite{MSCthesisvissearch}. 

\begin{figure}[htbp]
\centering
   \subfigure[\label{genworkflow} Bottle Category]{%
      \includegraphics[scale=0.28]{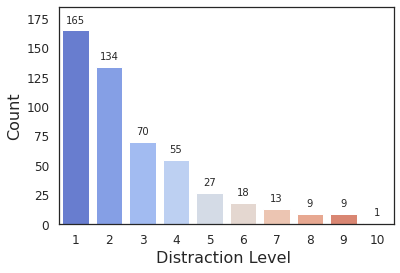}}
\hspace{1em}%
   \subfigure[\label{pyramidprocess} Car Category]{%
      \includegraphics[scale=0.28]{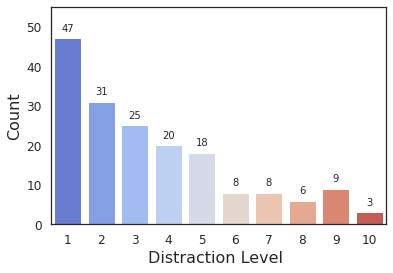}}
\hfill
   \subfigure[\label{pyramidprocess} Bowl Category]{%
      \includegraphics[scale=0.28]{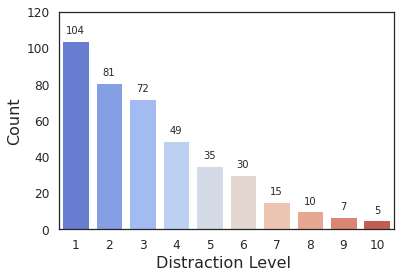}}
\caption{\label{workflow} The distribution of distraction levels in bottle, bowl, and car data. Distraction levels of 1 and 2 are the most prevalent, indicating that most distractors are fixated by one or two observers. A distraction level of more than 7 while searching for bottle and bowl, and a distraction level of more than 5 while searching for car is relatively rare.}
\label{fig:dist_level_distribu}
\end{figure}

\begin{figure}[p]
    \centering
\subfigure[]{
  \centering
  \includegraphics[width=0.22\textwidth]{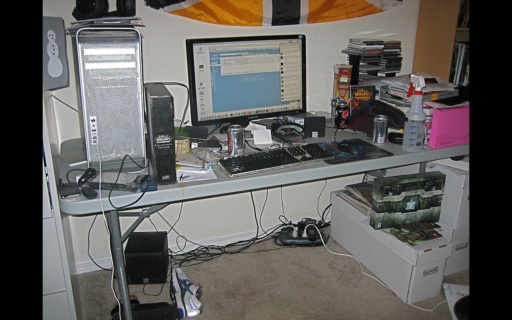}
}
\subfigure[]{
  \centering
  \includegraphics[width=0.22\textwidth]{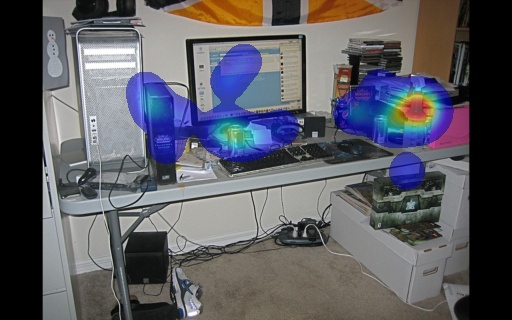}
}
\subfigure[]{
  \centering
  \includegraphics[width=0.22\textwidth]{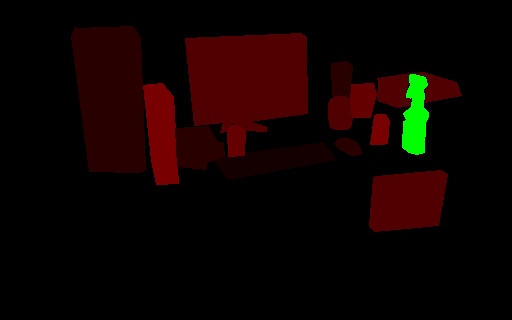}
}

\caption{\label{workflow} (a) Search image stimuli; (b) ground-truth fixation density map; (c) RGB-mapped object segmentation. Green indicates the target (bottle), and red objects are distractors. More intense red colors correspond to higher distraction levels.}
\label{fig:dist_level}
\end{figure}

To understand what percentage of distractors are highly distracting, we assign a distraction level to each distractor. We define distraction level of a distractor, as the number of participants who have fixated on it. As we have 10 participants in COCOSearch-18, the distraction levels vary between 0 and 10. A score of 0 means no distraction (no fixation), and 10 means a high distraction. Figure \ref{fig:dist_level_distribu} shows the histogram of distraction levels for fixated distractors in three target categories. As expected, most distractors are fixated by 1-3 observers. There are few cases where all 10 observers fixated at the same distractor. Figure \ref{fig:dist_level} shows a segmented image mapped to RGB color space based on the distraction levels.

\begin{figure}[p]
\centering
\includegraphics[width=0.5\textwidth]{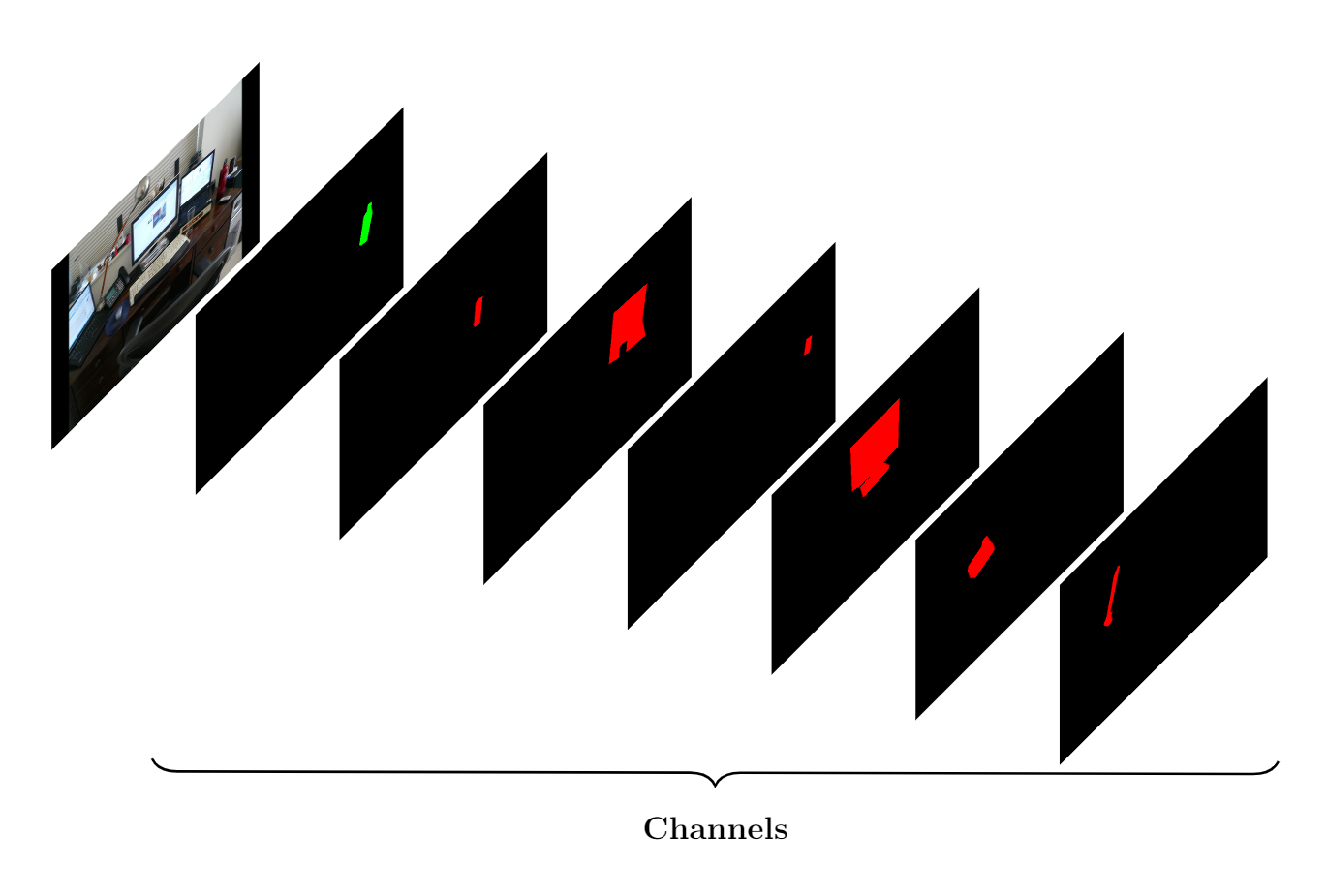}
\caption{The RGB-mapped segmentation masks used as input to Mask-RCNN pipeline. }
\label{fig:channel_s}
\end{figure}

Using our annotation labels, we create the ground truth instance segmentation of targets and fixated distractors such that each object forms a separate channel in a one-hot encoded format, as in figure \ref{fig:channel_s}. In this figure, the first channel contains the target object and the other channels contain the distractors (the number of distractors vary for each image, but they are limited to 20 instances). Considering the complexity of Mask-RCNN and for a more efficient use of data, we do not divide dataset into train-validation-test sets. Instead, we combine the training and validation splits of COCOSearch-18 and use a 5-fold cross validation method to train and evaluate the model on the whole data.

\subsection{Training}

\begin{figure}[p]
\centering
\includegraphics[scale=0.8]{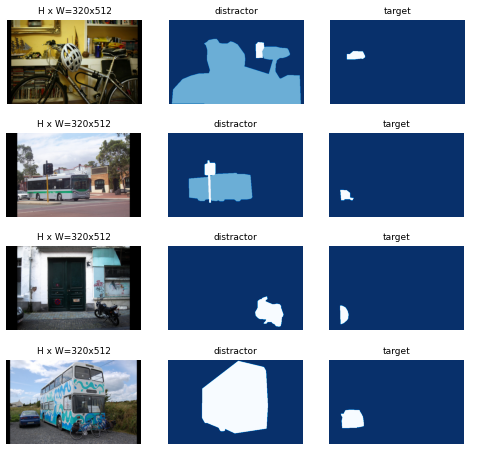}
\caption{The segmentation masks for each image are categorized into `target' and `distractor' classes in Mask-RCNN pipeline.}
\label{fig:showmask}
\end{figure}

Our segmentation problem is composed of 2 object classes `target' and `distractor' along with the default background class. The segmentation masks shown in figure \ref{fig:channel_s} are then reconfigured in MASK-RCNN pipeline based on their segmentation class, as shown in figure \ref{fig:showmask}. 

We use ResNet101 with strides 4, 8, 16, 32, and 64, as the backbone feature extractor. We initialize MASK-RCNN with pre-trained weights on MS COCO dataset. Then we perform a two-stage fine-tuning with progressively lower initial learning rates. The momentum and weight decay are set to 0.9, and 0.0001 for both stages. In the first stage, we freeze all layers, except the head branches. The head branches are composed of RPN, classifier and the mask heads of the network. We train these head layers for 10 epochs with a learning rate of 0.002. In the second stage, we fine-tune all layers with a learning rate of 0.001. In both stages, we also augment the data with horizontal or vertical flipping, each being chosen with 50\% probability.

We use a batch size of 1 for our training and validation. The network is able to detect maximum 100 object instances. The input images are of 320 × 512 dimension, and are transformed to square shape of 512 × 512 dimension. There are 5 different losses that are jointly minimized during training, in an end-to-end fashion. These losses are: RPN classification loss, RPN bounding box regression loss, M-RCNN classification loss, M-RCNN bounding box regression loss, and M-RCNN instance segmentation loss. All of these losses have equal weights (importance) and are calculated for both training and validation sets, at each epoch. During training the model tries to minimize the summation of all these losses, as seen in equation \ref{eq:loss}.

\begin{equation}
\begin{aligned}
Loss = rpn\textunderscore class\textunderscore loss + rpn\textunderscore bbox\textunderscore loss \\ + mrcnn\textunderscore class\textunderscore loss + mrcnn\textunderscore bbox\textunderscore loss \\ +  mrcnn\textunderscore mask\textunderscore loss \\
\label{eq:loss}
\end{aligned}
\end{equation}

At the end of each epoch, we calculate mean average precision (MAP), mean average recall (MAR), and f1-score at IoU threshold of 0.5 for the validation set. We save the weights of the model if the f1-score of validation set has increased compared to the previous epoch. To compute MAP, we first compute precision and recall values for each image. Precision or positive predictive value indicates what ratio of the predicted objects are true positives, i.e. they correspond to actual ground truth objects (A true positive happens when the predicted object class is correct and the predicted mask has an overlap of more than IoU threshold of 0.5 with the ground truth mask). Precision formula is shown in equation \ref{eq:precision}. Recall or sensitivity, as shown in equation \ref{eq:recall}, indicates what ratio of all the ground truth objects are correctly predicted by the model. 

\begin{equation}
\begin{aligned}
precision = \frac{TP}{TP + FP}
\label{eq:precision}
\end{aligned}
\end{equation}

\begin{equation}
\begin{aligned}
recall = \frac{TP}{TP + FN}
\label{eq:recall}
\end{aligned}
\end{equation}

Average precision (AP) can be calculated for each object class, as the area under the precision-recall curve for that class. Then mean average precision (MAP), is the mean of AP over all object classes (target and distractor). For mean average recall (MAR), we compute recall values for IoU threshold of 0.5, and take the average over images and object classes. When our M-RCNN has a high recall but low precision, it predicts most of the targets and distractors correctly but it also has many false positives and predicts many background objects as distractors (or sometimes targets). While, when it has high precision but low recall, then most of the objects it detects as distractors or target are accurate, but it misses some of the distractor and target objects. F1-score, which is the harmonic mean of precision and recall, finds a balance between these two metrics. F1-score is calculated using equation \ref{eq:F1score}. A high f1-score indicates a high precision and recall. 

\begin{equation}
\begin{aligned}
F1-score = \frac{2 (MAP_{50}.MAR_{50})}{MAP_{50} + MAR_{50}} = \frac{TP}{TP + \frac{1}{2}(FP + FN)}
\label{eq:F1score}
\end{aligned}
\end{equation}

\subsection{Experiments and Analysis}

As we need to train a separate Mask-RCNN for each target category and the amount of data for each category is comparatively small, we do not separate a test set from our data. Instead, we run 5-fold cross-validation, dividing data into 5 folds, using one different fold as the validation data at each round, and training the model on the rest. In each cross validation round, after the training ends, we compute the MAP and MAR of the best check-pointed model on the validation data of that round. After 5 rounds of cross-validation, we then compute the average MAP, MAR, and f1-score over all 5 validation folds. We repeat this cross-validation for 5 independent runs (25 validation rounds in total) and report the mean and standard deviations of the results in table \ref{table:target_dist_results}. This method helps us to validate the model's performance on all data and average out the variations in the difficulty of images among different validation sets. We also consider a baseline model, that is a MASK-RCNN initialized with pre-trained weights on MS COCO dataset without any further training. We compare our trained models for each of the three target categories with this baseline model in table \ref{table:target_dist_results}. The Baseline model has a low performance with an average MAP\textsubscript{50} of 0.005, and MAR\textsubscript{50} of 0.087 and f1-score of 0.010, over the three target categories. This shows that our training improves the model performance considerably. Our trained model has an average MAP\textsubscript{50} of 0.570 , $114\times$ better than the baseline, MAR\textsubscript{50} of 0.726, $8\times$ better than the baseline, and f1-score of 0.637, $6\times$ better than the baseline. Here, the baseline model should be better than chance (a random model), as it is pre-trained on MS COCO dataset to detect objects. Yet, it has much worse scores compared to our trained model. This shows that our model performs much better than chance in predicting targets and distractors.  

\begin{figure}[htbp]
    \centering
\subfigure[bottle]{
  \centering
  \includegraphics[width=0.22\textwidth]{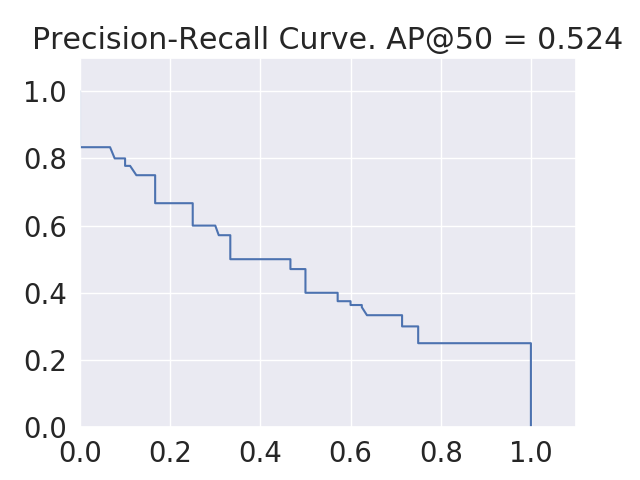}
  \label{fig:fig3}
}
\subfigure[bowl]{
  \centering
  \includegraphics[width=0.22\textwidth]{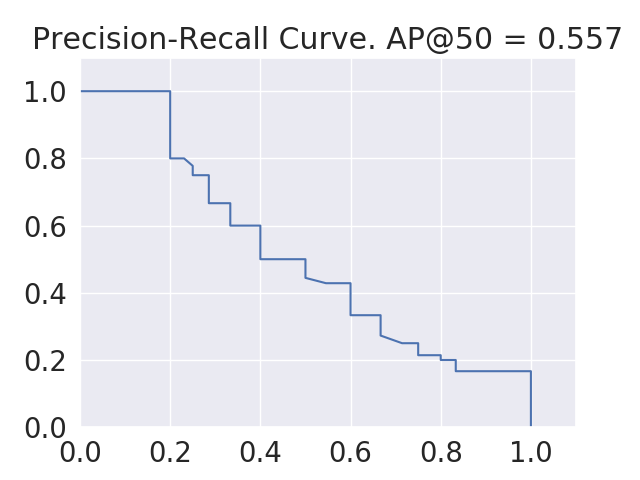}
  \label{fig:fig4}
}
\subfigure[car]{
  \centering
  \includegraphics[width=0.22\textwidth]{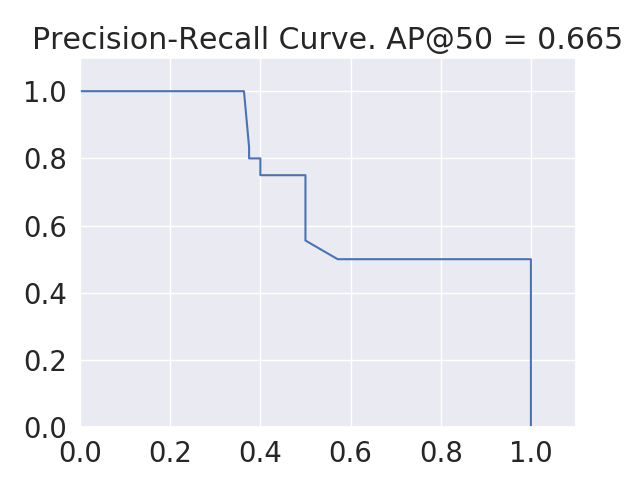}
  \label{fig:fig5}
}
\caption{\label{workflow} The PR curves of the three MASK-RCNN models, trained on each of bottle, bowl, and car target categories. The curves contain the precision and recall values for one full run of 5-fold cross validation.}
\label{fig:pres-recall}
\end{figure}

In figure \ref{fig:pres-recall}, we show sample precision-recall curves for each of the three models of bottle, bowl, and car target categories. Each curve contains the precision and recall values on the validation data of a full run of 5-fold cross validation over the whole dataset. Clearly, the highest area under the PR curve belongs to car, then bowl, and finally bottle target category. This is in accordance with the MAP\textsubscript{50} scores we obtained for these categories, i.e. car has the largest and bottle has the lowest value.

\begin{table}[htbp]
\captionsetup{justification=centering}
\caption{The average performance of the trained Mask-RCNN and Baseline model on COCOSearch18 dataset for bottle, bowl, and car target categories. The values are averaged over 5 independent runs of 5-fold cross validation.} 



\makebox[0.5\textwidth][c]{
    \begin{tabular}{cccccc}
\toprule
\multirow{2}{*}{\textbf{Category}} & \multicolumn{2}{c}{\multirow{2}{*}{\textbf{Model}}}& \multicolumn{3}{c}{ \textbf{Metrics}}  \\

\cmidrule(lr){4-6}
&  &  & \textbf{MAP\textsubscript{0.5}} & \textbf{MAR\textsubscript{0.5}} & \textbf{F1-score}\\
\midrule

\multirow{4}{*}{\textbf{Bottle}} & \multirow{2}{*}{\textbf{Trained MR-CNN}} & \textbf{$\mu$} &0.513	&0.724	&0.600\\
&  & \textbf{$\sigma$} &0.007 &	0.015 &	0.009\\

\cmidrule(lr){2-6}
 & \multirow{1}{*}{\textbf{Baseline MR-CNN}}& \textbf{$\mu$}  &0.004&	0.150&	0.008\\
\midrule

\multirow{4}{*}{\textbf{Bowl}} & \multirow{2}{*}{\textbf{Trained MR-CNN}} & \textbf{$\mu$} &0.547&	0.715&	0.620\\
&  & \textbf{$\sigma$} & 0.008	&0.002 &	0.005\\
\cmidrule(lr){2-6}
 & \multirow{1}{*}{\textbf{Baseline MR-CNN}} & \textbf{$\mu$}  &0.000	&0.030	&0.000\\
\midrule

\multirow{4}{*}{\textbf{Car}} & \multirow{2}{*}{\textbf{Trained MR-CNN}} & \textbf{$\mu$} &0.650	 &0.739 &	0.692\\
&  & \textbf{$\sigma$} &0.013&	0.016&	0.013\\
\cmidrule(lr){2-6}
 & \multirow{1}{*}{\textbf{Baseline MR-CNN}} & \textbf{$\mu$}&  0.012	&0.080	&0.022\\
\midrule

 \multirow{4}{*}{\textbf{All (Avg.)}} & \multirow{2}{*}{\textbf{Trained MR-CNN}} & \textbf{$\mu$} &0.570&	0.726&	0.637\\
&  & \textbf{$\sigma$} & 0.072	&0.012&	0.048\\
\cmidrule(lr){2-6}
 & \multirow{1}{*}{\textbf{Baseline MR-CNN}} & \textbf{$\mu$}&  0.005	&0.087	&0.010\\
\midrule

\end{tabular}
}
\label{table:target_dist_results}

\end{table}%

\begin{figure*}[htbp]
    \centering
\subfigure[bottle]{
  \centering
  \includegraphics[width=0.45\textwidth]{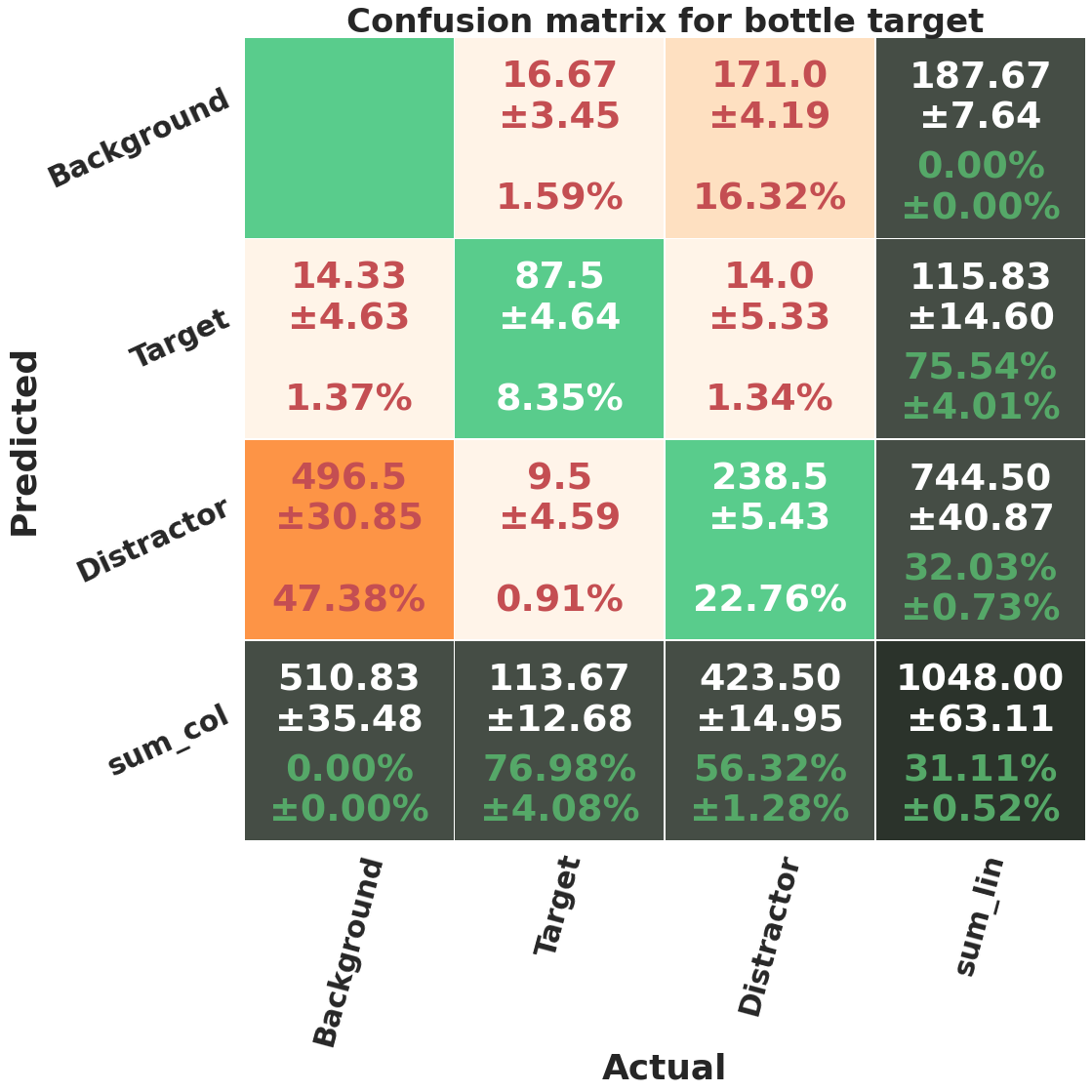}
}
\hspace{0.1 cm}
\subfigure[bowl]{
  \centering
  \includegraphics[width=0.45\textwidth]{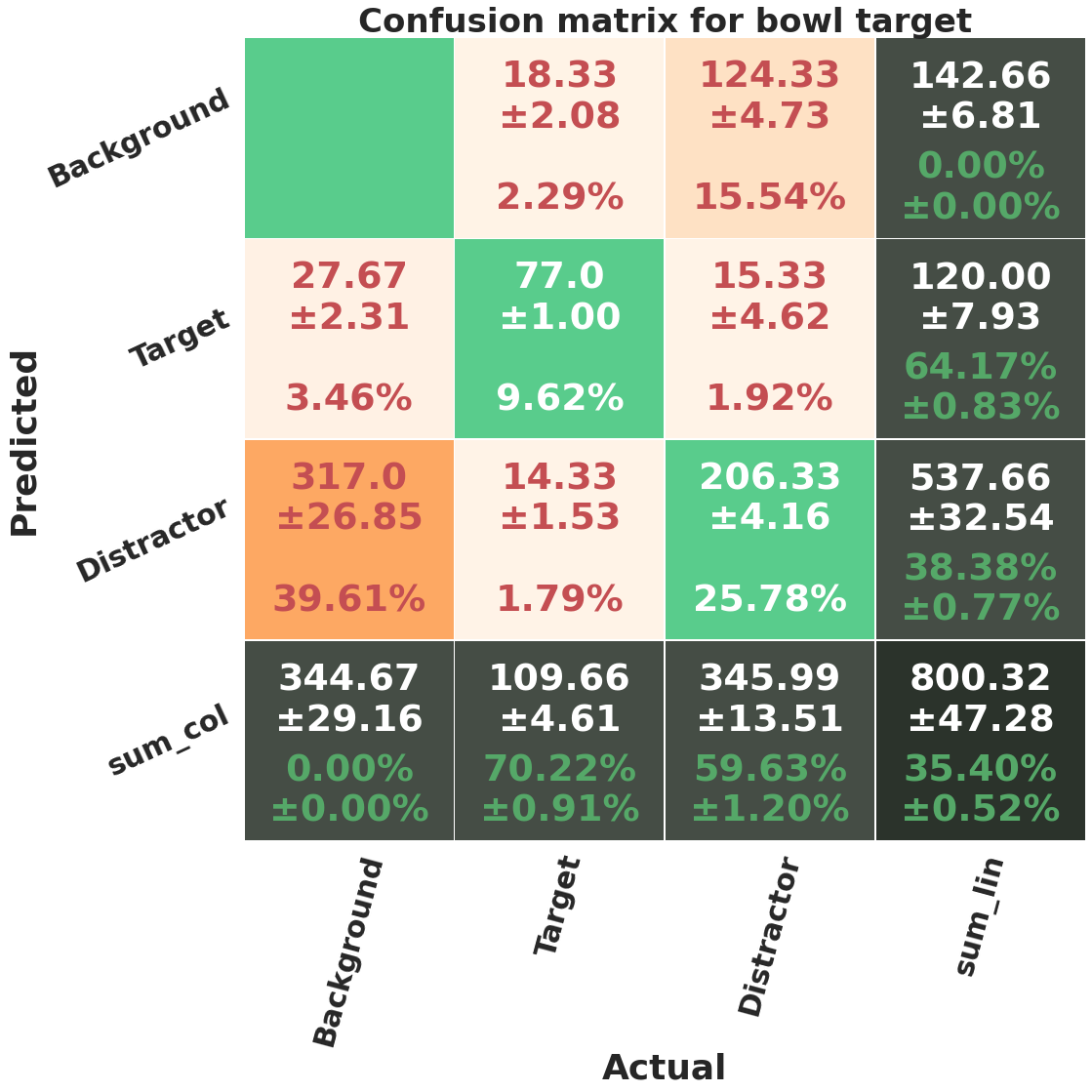}
}
\hspace{0.1 cm}
\subfigure[car]{
  \centering
  \includegraphics[width=0.45\textwidth]{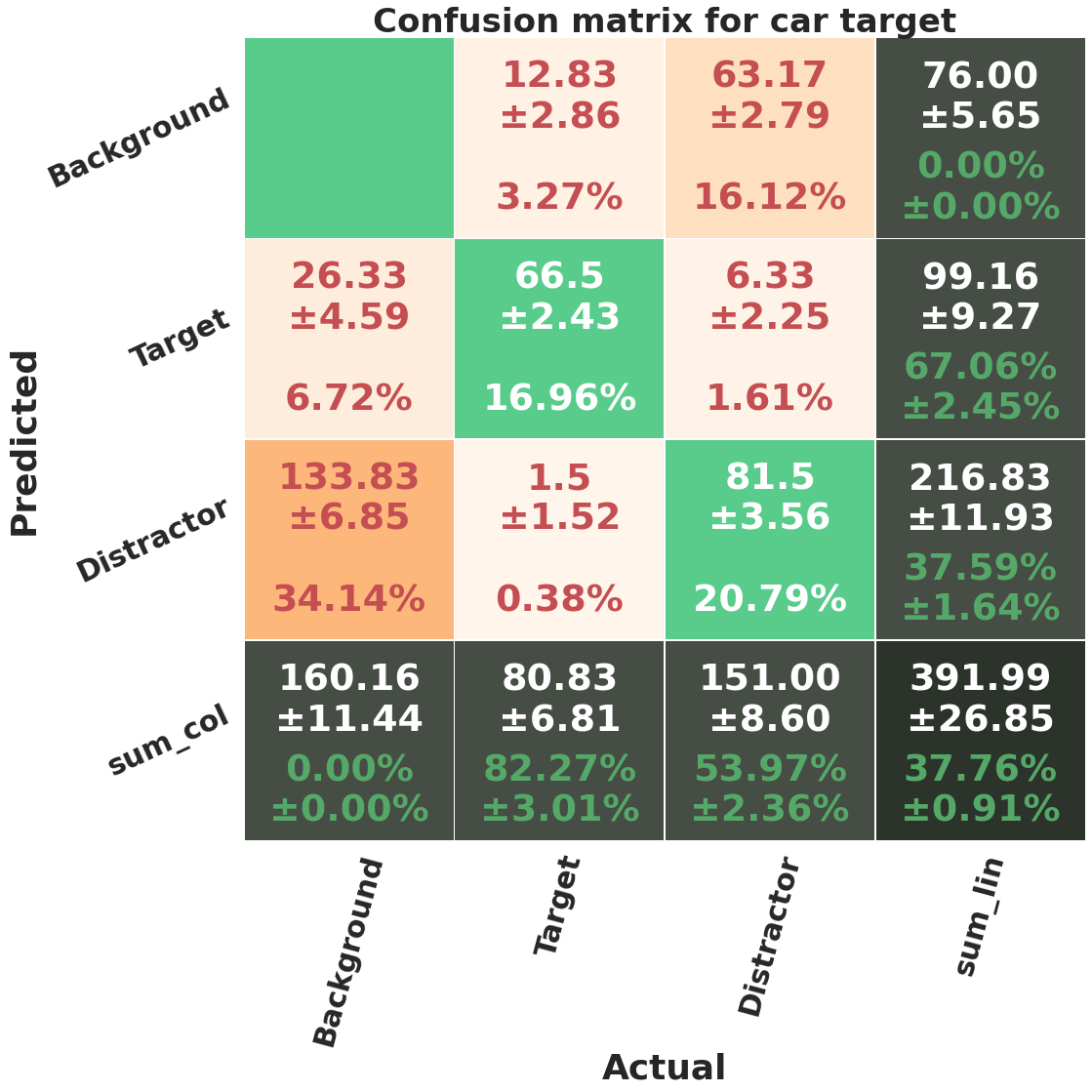}}

\caption{\label{workflow} The confusion matrices for bottle, bowl, and car categories, averaged over 5 independent full runs of 5-fold cross validation (25 rounds in total). The last columns and rows show the total counts of the previous columns/rows.}
\label{fig:cm_}
\end{figure*}

In figure \ref{fig:cm_}, we show the confusion matrices of each target category averaged over 25 runs (5 independent runs of 5 fold-cross validation). A prediction is considered true positive if the predicted and ground truth bounding boxes have an IoU of more than 0.5, and belong to the same object class (target/distractor). The ground truth objects that do not have an IoU of more than 0.5 with any of the predicted objects are predicted as background, and the predicted objects that do not pass the overlap threshold with any ground truth object has background label as their ground-truth. The accuracy of each class (target and distractor), which is computed using equation \ref{eq:acc}, can be seen in the last row of the matrices. The mean accuracy percentage and its standard deviation are shown in green font. We report these accuracies in table \ref{table:target-dist-acc}. 

\begin{equation}
\begin{aligned}
Accuracy = \frac{TP + TN}{TP + TN + FP + FN}
\label{eq:acc}
\end{aligned}
\end{equation}

The highest target detection accuracy belongs to car category, which can be related to the higher visibility and bigger size of target objects, and bowl category has the highest distractor detection accuracy. It can be seen in confusion matrices, that a significant number of background objects, $\approx40\%$, are predicted as distractors by the network. We hypothesise that if more observers were involved in the fixation data of COCO-Search18 dataset (which are currently only 10 people), we would expect that some of the other background objects appear as distractors. It is possible that by increasing the number of observers, the percentage of false positive for distractor class decrease. Also, $\approx 16\%$ of distractors are not predicted by the model, we will see later that these are mostly the less distracting objects. 

\begin{table}[htbp]
\captionsetup{justification=centering}
\caption{The average accuracy of the trained Mask-RCNN for bottle, bowl, and car target categories obtained from confusion matrix. The values are averaged over 5 independent runs of 5-fold cross validation.}

\makebox[0.5\textwidth][c]{
    \begin{tabular}{ccc}
\toprule
\multirow{2}{*}{\textbf{Category}} & \multicolumn{2}{c}{\textbf{Accuracy}}  \\
\cmidrule(lr){2-3}

 & \textbf{Target} & \textbf{Distractor}\\
\midrule

\textbf{Bottle} & $\textbf{76.98\%} \pm 4.08\% $& $\textbf{56.32\%} \pm 1.28\% $\\
\midrule

\textbf{Bowl} & $\textbf{70.22\%} \pm 0.91\% $ & $\textbf{59.63\%} \pm 1.20\% $\\
\midrule

\textbf{Car} & $\textbf{82.27\%} \pm 3.01\%$ & $\textbf{53.97\%} \pm 2.36\%$ \\
\midrule

\end{tabular}
}
\label{table:target-dist-acc}
\end{table}%

\begin{figure*}[htbp]
\centering
\subfigure{
  \centering
  \includegraphics[scale = 0.45]{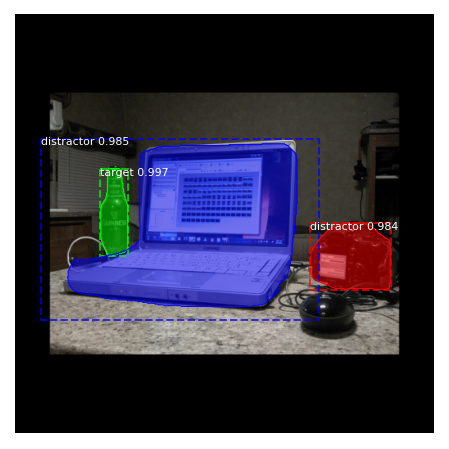}
}
\subfigure{
  \centering
  \includegraphics[scale = 0.45]{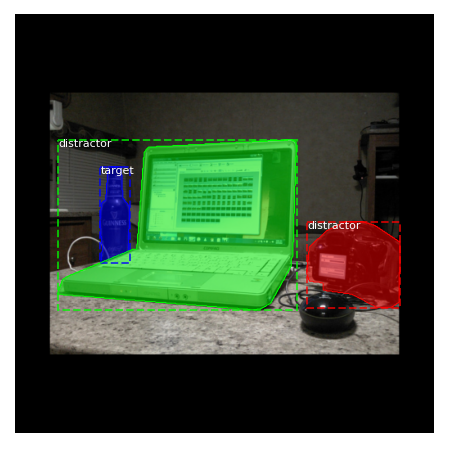}
  \label{fig:fig2}
}
\subfigure{
  \centering
  \includegraphics[scale = 0.45]{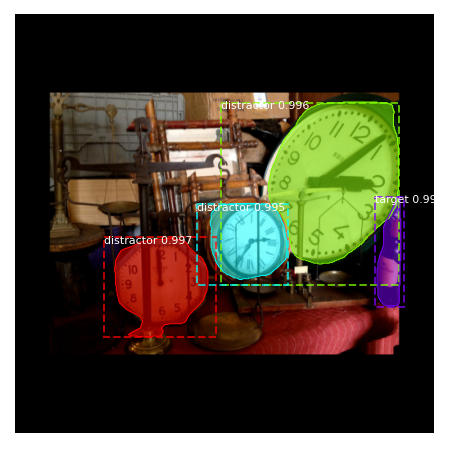}
  \label{fig:fig3}
}
\subfigure{
  \centering
  \includegraphics[scale = 0.45]{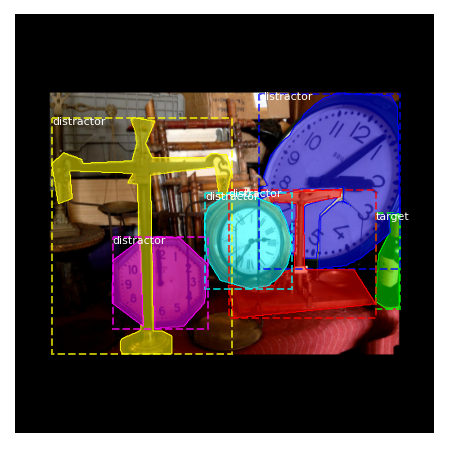}
  \label{fig:fig4}
}

\subfigure{
  \centering
  \includegraphics[scale = 0.45]{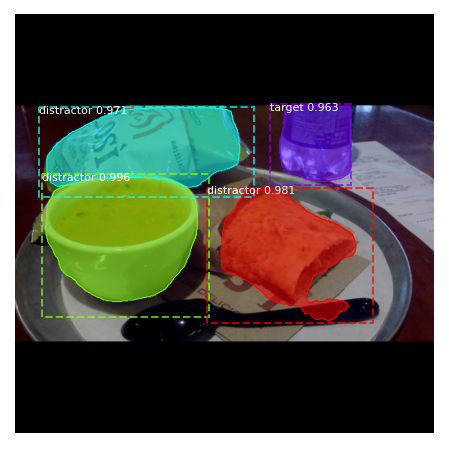}
  \label{fig:fig5}
}
\subfigure{
  \centering
  \includegraphics[scale = 0.45]{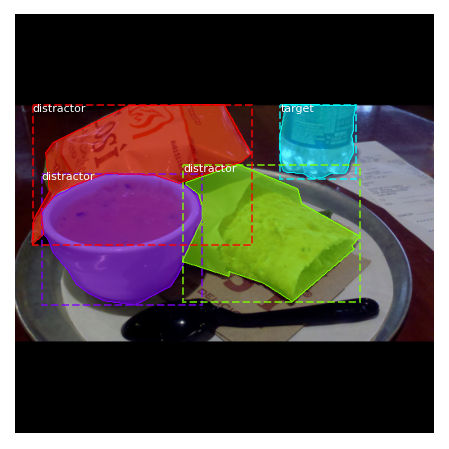}
  \label{fig:fig6}
}

\caption{Sample results of network predictions on validation data for `Bottle' target category. The left column contains the predictions. The right column contains the ground-truth segmentations.}

\label{fig:bottle_category}
\end{figure*}

\begin{figure*}[htbp]
\centering
\subfigure{
  \centering
  \includegraphics[scale = 0.45]{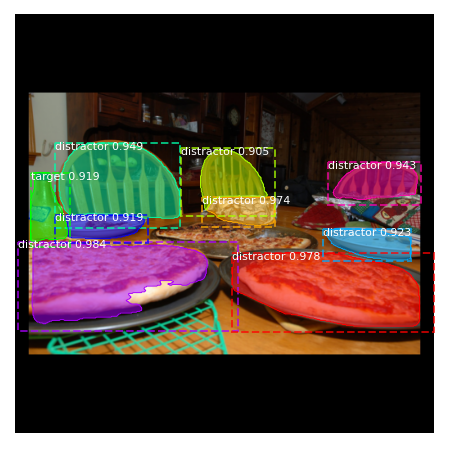}
}
\subfigure{
  \centering
  \includegraphics[scale = 0.45]{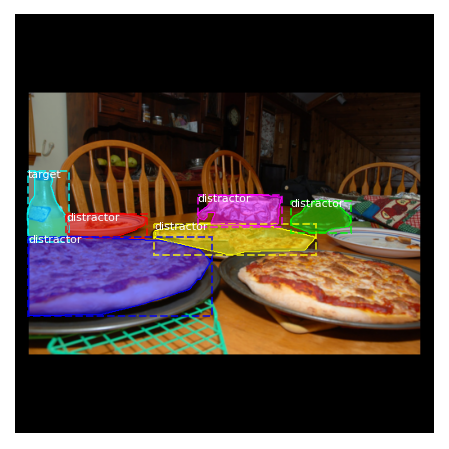}
  \label{fig:fig2}
}
\subfigure{
  \centering
  \includegraphics[scale = 0.45]{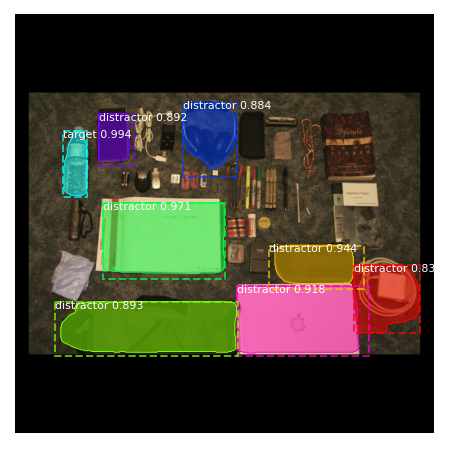}
  \label{fig:fig3}
}
\subfigure{
  \centering
  \includegraphics[scale = 0.45]{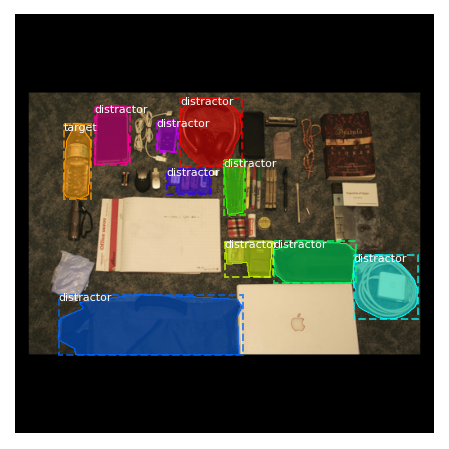}
  \label{fig:fig4}
}

\subfigure{
  \centering
  \includegraphics[scale = 0.45]{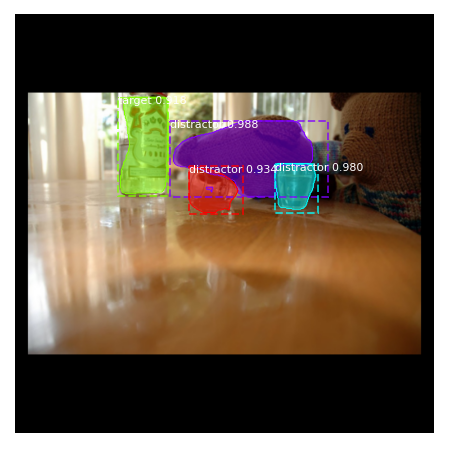}
  \label{fig:fig5}
}
\subfigure{
  \centering
  \includegraphics[scale = 0.45]{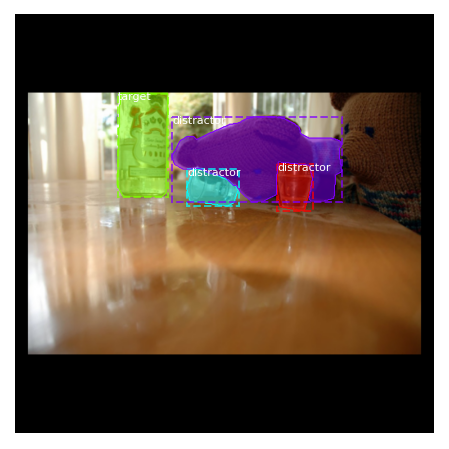}
  \label{fig:fig6}
}

\caption{Sample results of network predictions on validation data for `Bottle' target category. The left column contains the predictions. The right column contains the ground-truth segmentations.}

\label{fig:bottle_category_2}
\end{figure*}

\begin{figure*}[htbp]
\centering
\subfigure{
  \centering
  \includegraphics[scale = 0.45]{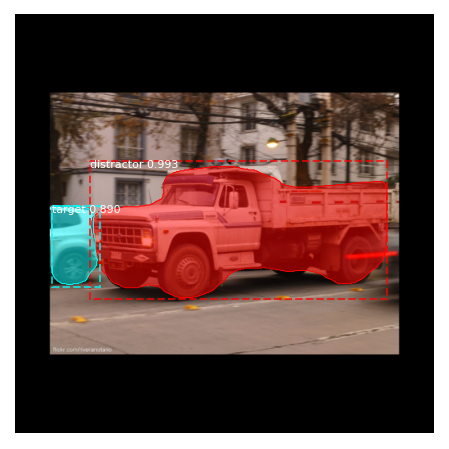}
}
\subfigure{
  \centering
  \includegraphics[scale = 0.45]{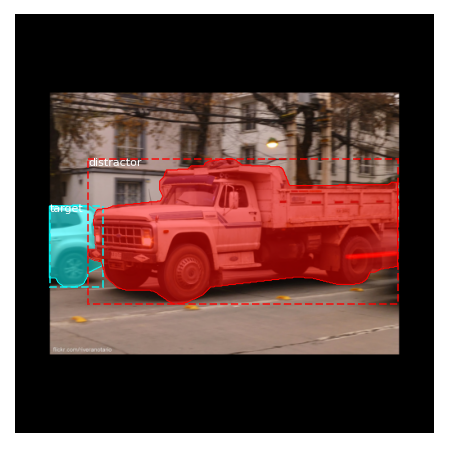}
  \label{fig:fig2}
}
\subfigure{
  \centering
  \includegraphics[scale = 0.45]{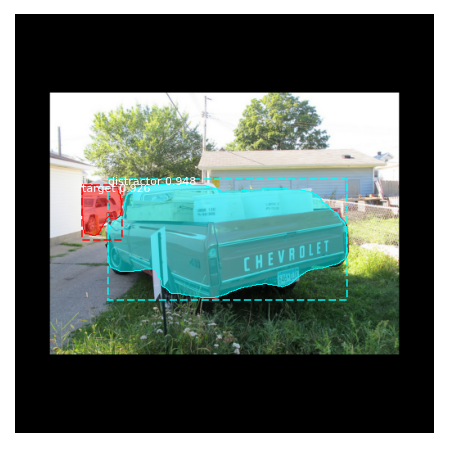}
  \label{fig:fig3}
}
\subfigure{
  \centering
  \includegraphics[scale = 0.45]{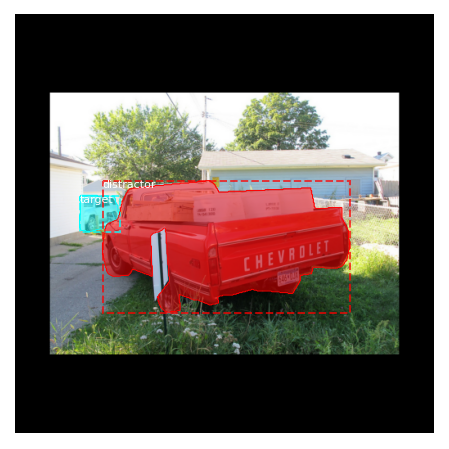}
  \label{fig:fig4}
}

\subfigure{
  \centering
  \includegraphics[scale = 0.45]{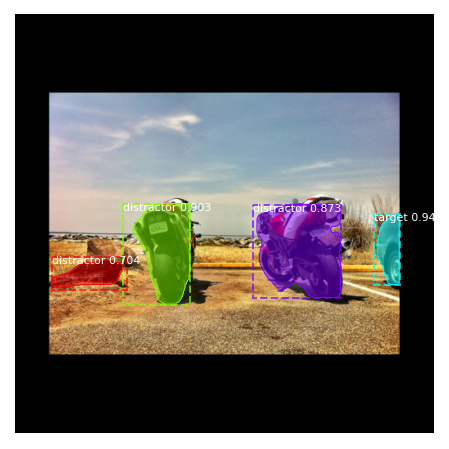}
  \label{fig:fig5}
}
\subfigure{
  \centering
  \includegraphics[scale = 0.45]{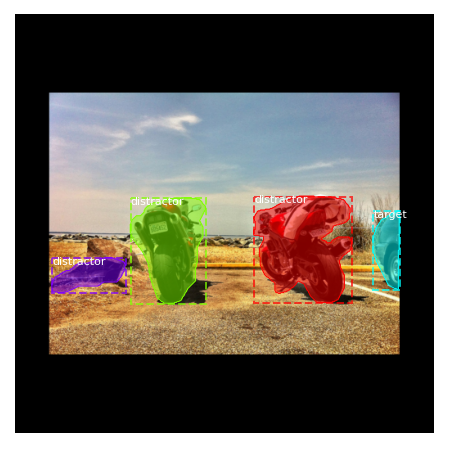}
  \label{fig:fig6}
}

\caption{Sample results of network predictions on validation data for `Car' target category. The left column contains the predictions. The right column contains the ground-truth segmentations.}

\label{fig:car_category}
\end{figure*}

\begin{figure*}[htbp]
\centering
\subfigure{
  \centering
  \includegraphics[scale = 0.45]{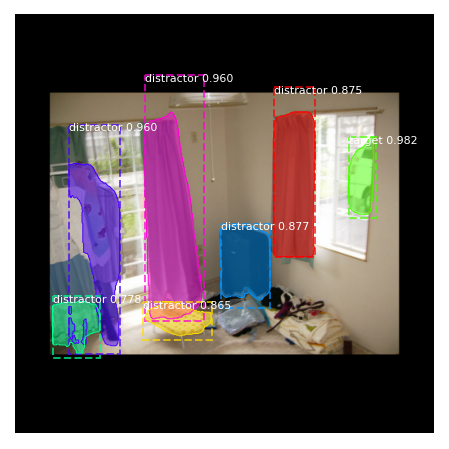}
}
\subfigure{
  \centering
  \includegraphics[scale = 0.45]{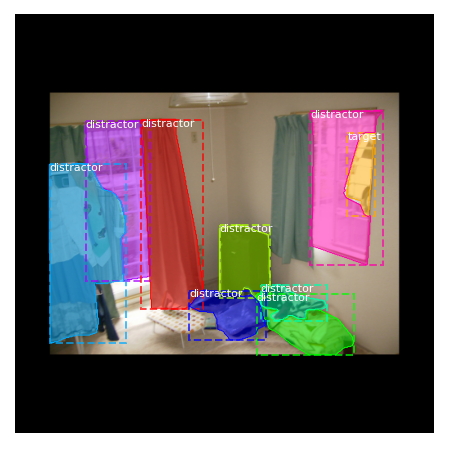}
  \label{fig:fig2}
}
\subfigure{
  \centering
  \includegraphics[scale = 0.45]{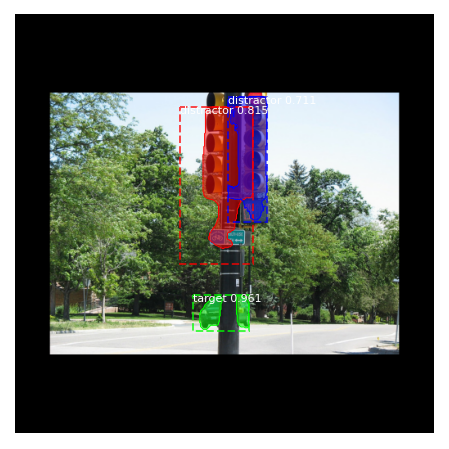}
  \label{fig:fig3}
}
\subfigure{
  \centering
  \includegraphics[scale = 0.45]{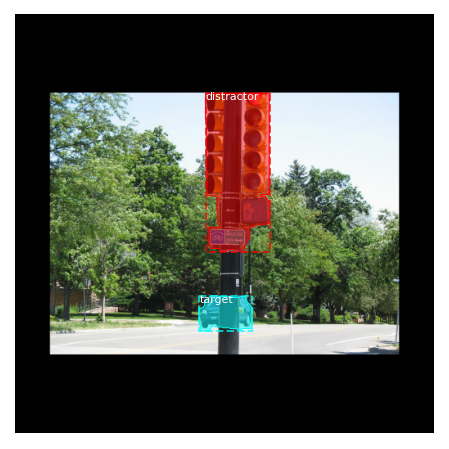}
  \label{fig:fig4}
}

\subfigure{
  \centering
  \includegraphics[scale = 0.45]{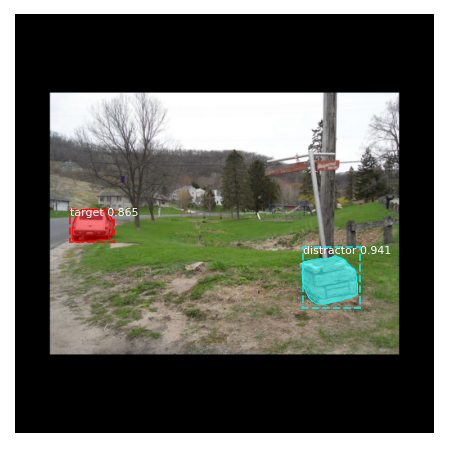}
  \label{fig:fig5}
}
\subfigure{
  \centering
  \includegraphics[scale = 0.45]{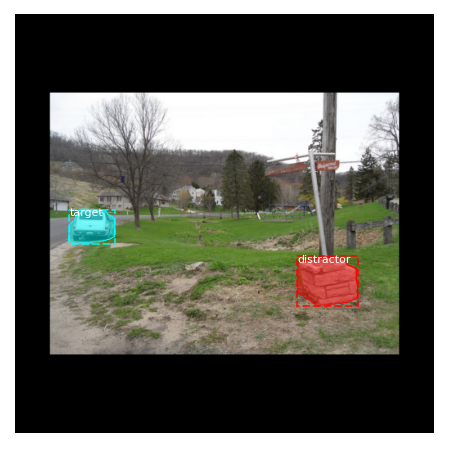}
  \label{fig:fig6}
}

\caption{Sample results of network predictions on validation data for `Car' target category. The left column contains the predictions. The right column contains the ground-truth segmentations.}

\label{fig:car_category_2}
\end{figure*}

\begin{figure*}[htbp]
\centering
\subfigure{
  \centering
  \includegraphics[scale = 0.45]{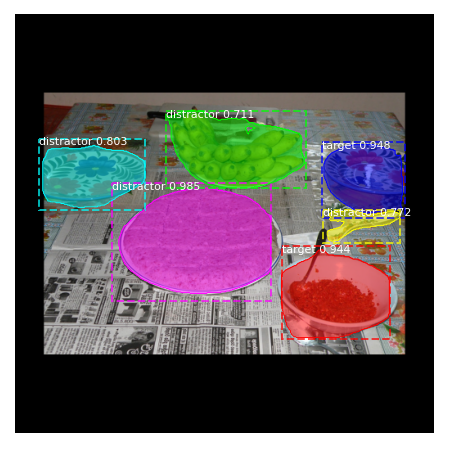}
}
\subfigure{
  \centering
  \includegraphics[scale = 0.45]{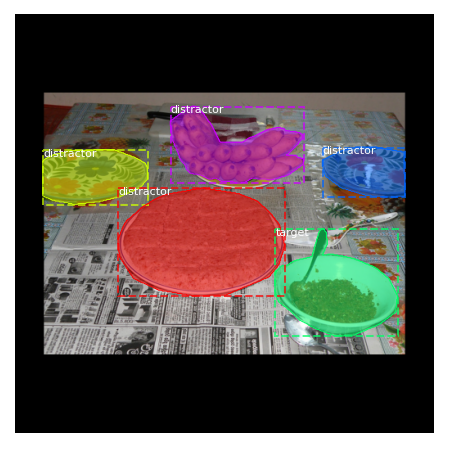}
  \label{fig:fig2}
}
\subfigure{
  \centering
  \includegraphics[scale = 0.45]{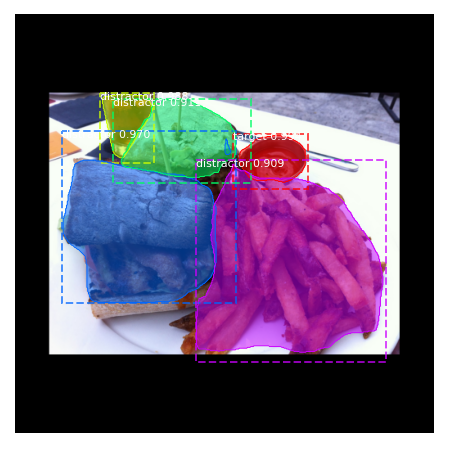}
  \label{fig:fig3}
}
\subfigure{
  \centering
  \includegraphics[scale = 0.45]{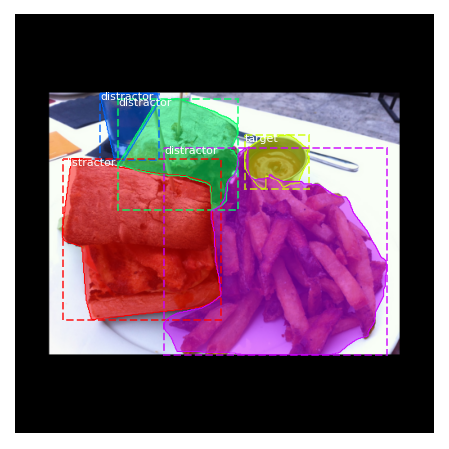}
  \label{fig:fig4}
}

\subfigure{
  \centering
  \includegraphics[scale = 0.45]{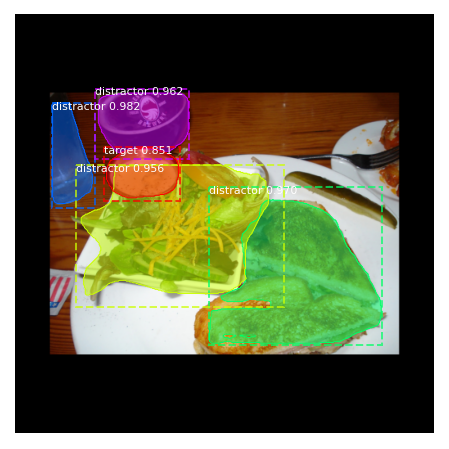}
  \label{fig:fig5}
}
\subfigure{
  \centering
  \includegraphics[scale = 0.45]{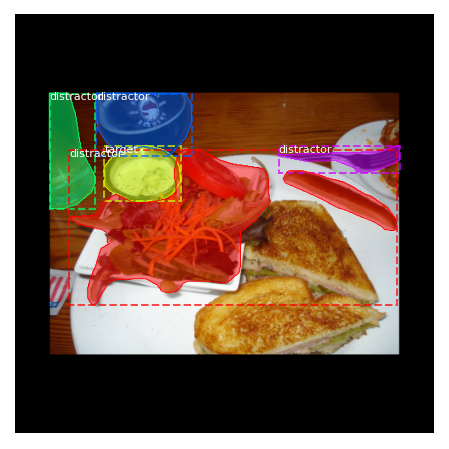}
  \label{fig:fig6}
}

\caption{Sample results of network predictions on validation data for `Bowl' target category. The left column contains the predictions. The right column contains the ground-truth segmentations.}

\label{fig:bowl_category}
\end{figure*}

\begin{figure*}[htbp]
\centering
\subfigure{
  \centering
  \includegraphics[scale = 0.45]{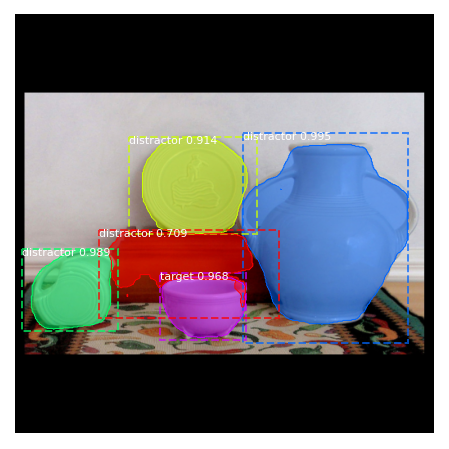}
}
\subfigure{
  \centering
  \includegraphics[scale = 0.45]{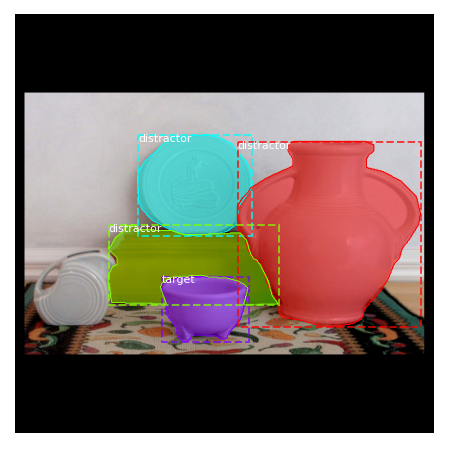}
  \label{fig:fig2}
}
\subfigure{
  \centering
  \includegraphics[scale = 0.45]{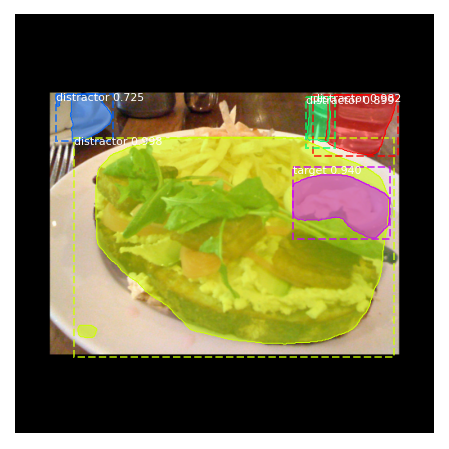}
  \label{fig:fig3}
}
\subfigure{
  \centering
  \includegraphics[scale = 0.45]{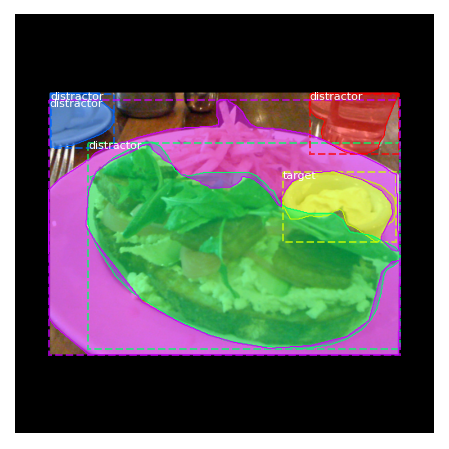}
  \label{fig:fig4}
}

\subfigure{
  \centering
  \includegraphics[scale = 0.45]{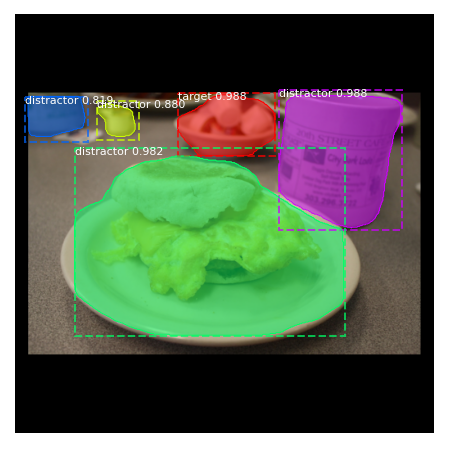}
  \label{fig:fig5}
}
\subfigure{
  \centering
  \includegraphics[scale = 0.45]{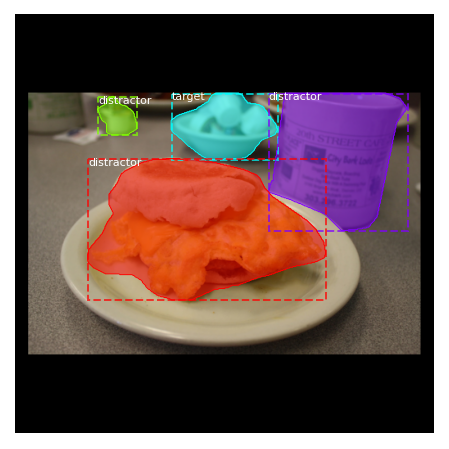}
  \label{fig:fig6}
}

\caption{Sample results of network predictions on validation data for `Bowl' target category. The left column contains the predictions. The right column contains the ground-truth segmentations.}

\label{fig:bowl_category_2}
\end{figure*}

Figures \ref{fig:bottle_category}, \ref{fig:bottle_category_2}, \ref{fig:car_category}, \ref{fig:car_category_2}, \ref{fig:bowl_category} and \ref{fig:bowl_category_2}, show sample predictions of M-RCNN along with their ground-truth segmentations. The network sometimes fails to detect the target mostly due to occlusion, blurriness, or a very small size. The network also outputs a confidence score/probability for each segmented object. In the generation of figures \ref{fig:bottle_category} to \ref{fig:bowl_category_2}, we have defined various thresholds on the distractor's confidence score to only output the top 3 or 4 predicted distractors in each image. Without defining these thresholds, the network shows all of the predicted distractors even those with low confidence scores, that do not even contain a real object. For most images a confidence threshold of ~90-95\% removes most false predictions. 
Our network segments and classifies many foreground objects that do not belong to the target category as distractors. However, factors such as similarity to the target, proximity to the target, proximity to the center of the image, and distractor's size, affects how distracting an object is. Defining a threshold on the confidence scores could be a way to separate the most distracting objects. 

\begin{figure*}[p]
    \centering
\subfigure[]{
  \centering
  \includegraphics[width=0.45\textwidth]{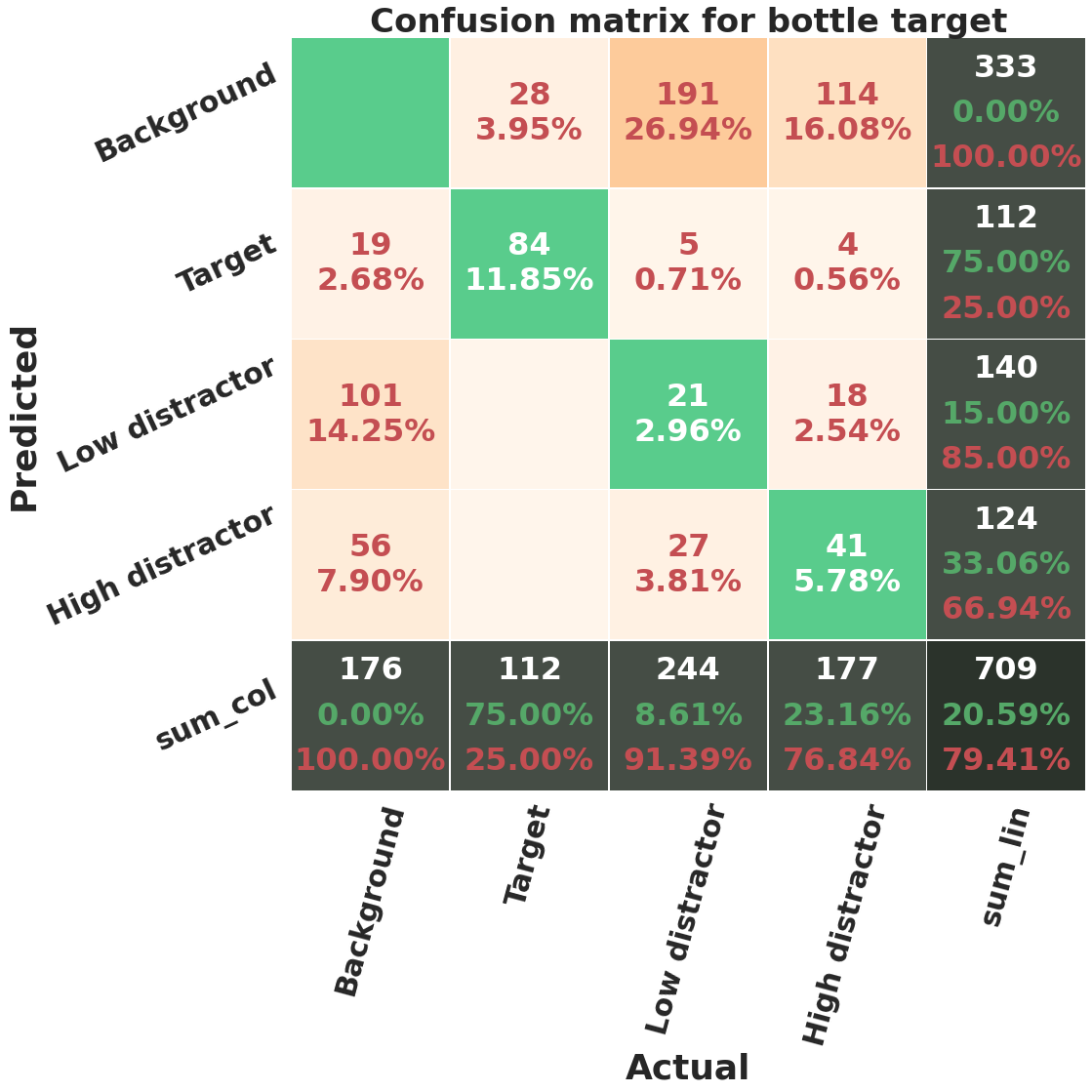}

}
\hspace{0.1 cm}
\subfigure[]{
  \centering
  \includegraphics[width=0.45\textwidth]{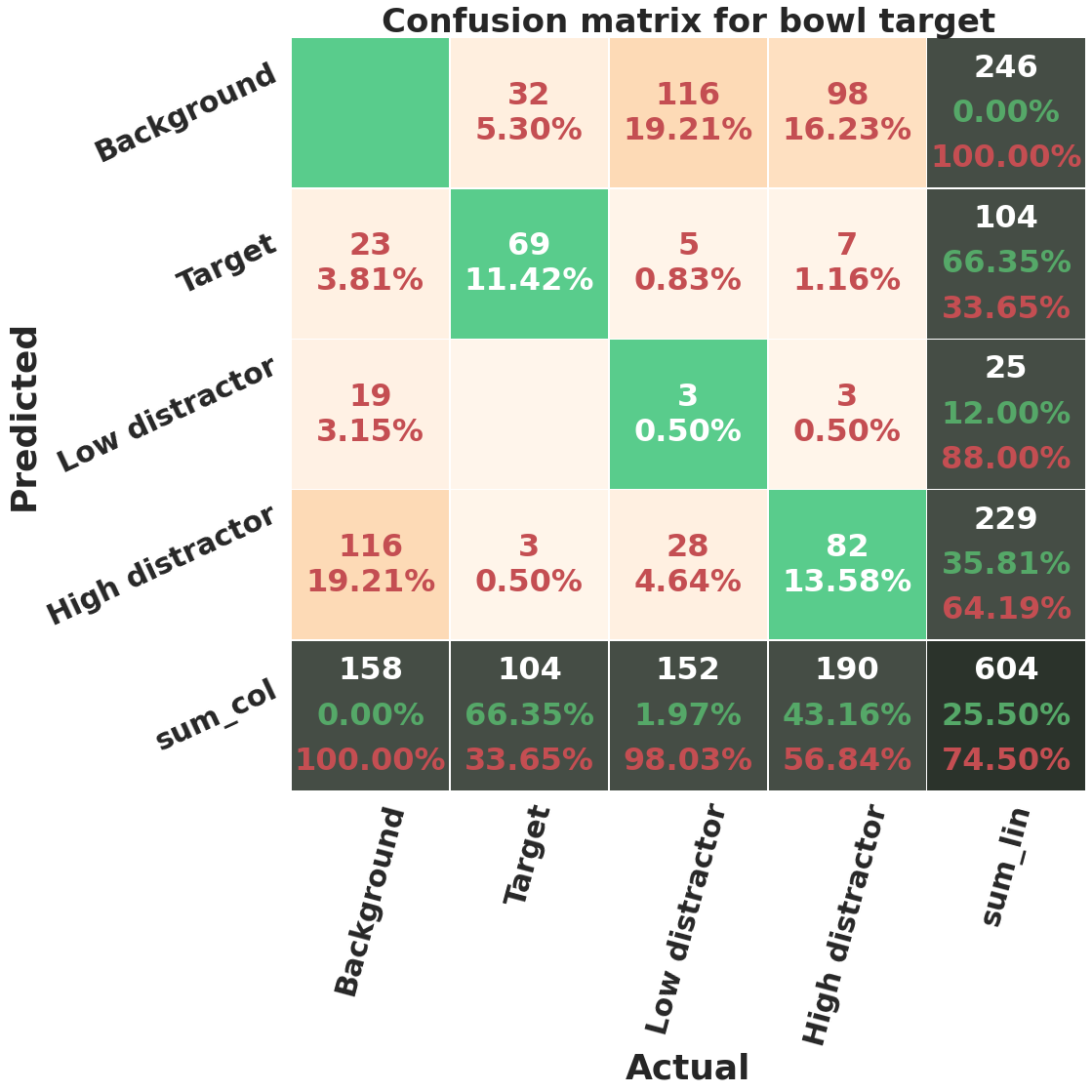}

}
\hspace{0.1 cm}
\subfigure[]{
  \centering
  \includegraphics[width=0.45\textwidth]{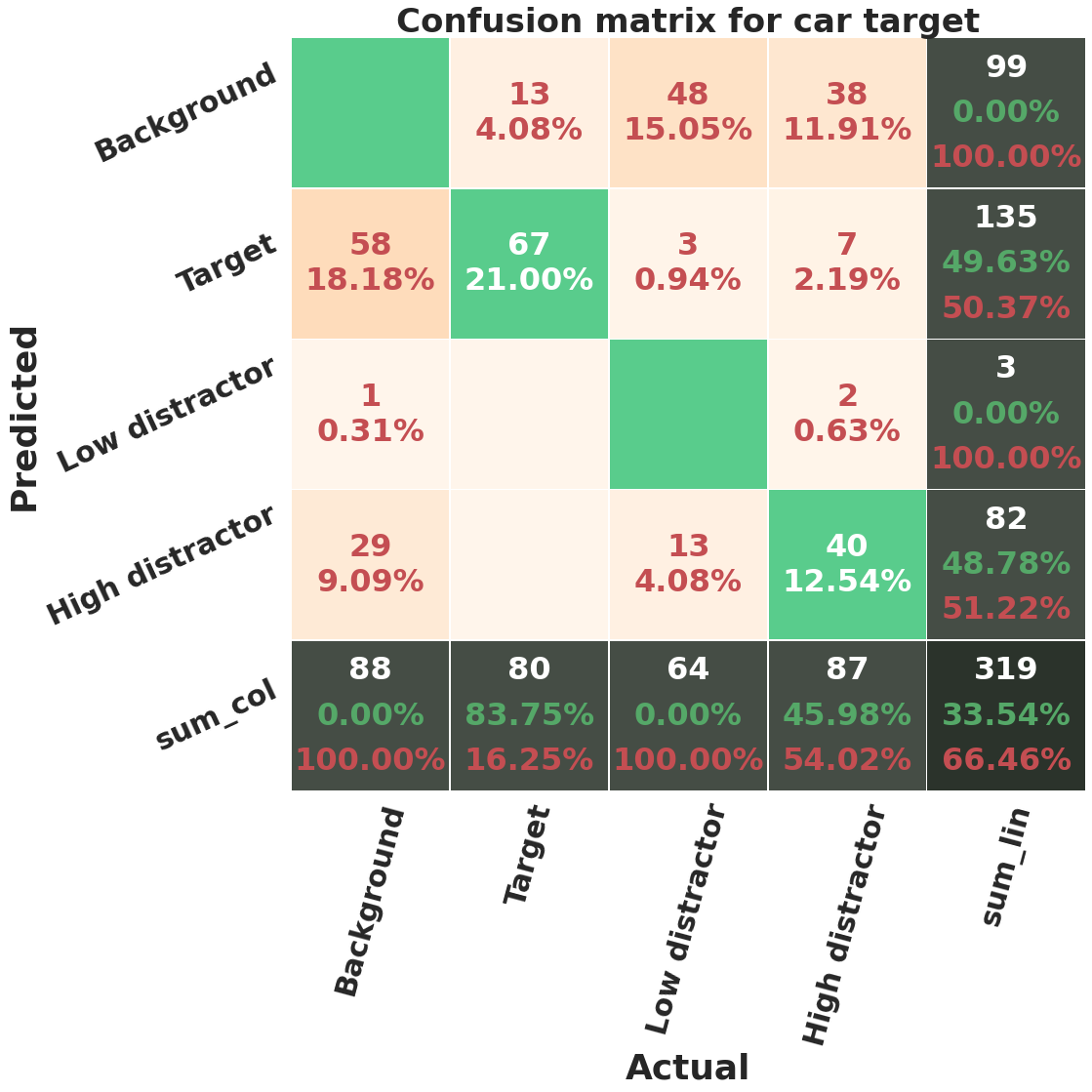}

}

\caption{\label{workflow} The confusion matrices for bottle, bowl, and car categories for 3-class classification: low-distractor, high-distractor, and target. Results are reported for one run of 5-fold cross validation (5 rounds in total).}
\label{fig:cm_3class}
\end{figure*}

\begin{figure*}
\centering

\subfigure{
  \centering
  \includegraphics[scale = 0.45]{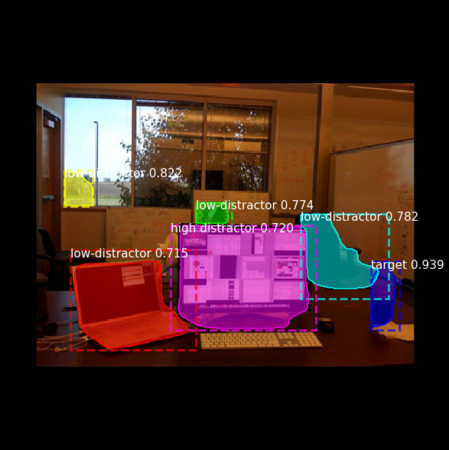}
  \label{fig:fig2}
}
\subfigure{
  \centering
  \includegraphics[scale = 0.45]{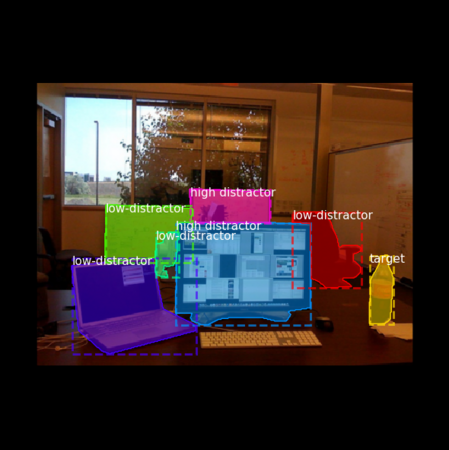}
  \label{fig:fig1}
}

\vfill
\vspace{4.00mm}

\subfigure{
  \centering
  \includegraphics[scale = 0.45]{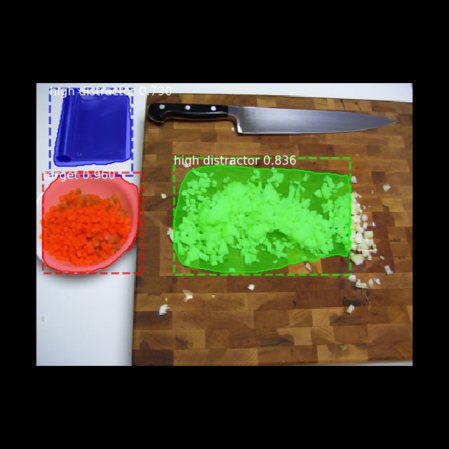}
  \label{fig:fig4}
}
\subfigure{
  \centering
  \includegraphics[scale = 0.45]{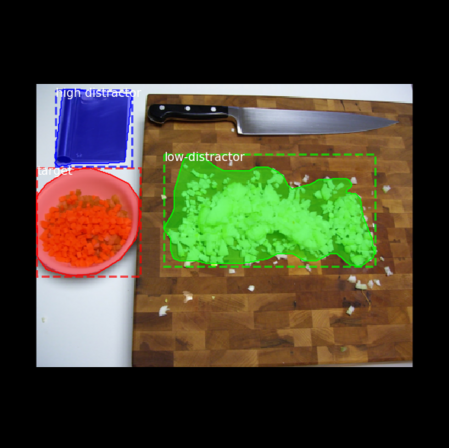}
  \label{fig:fig3}
}

\vfill
\vspace{4.00mm} 

\subfigure{
  \centering
  \includegraphics[scale = 0.45]{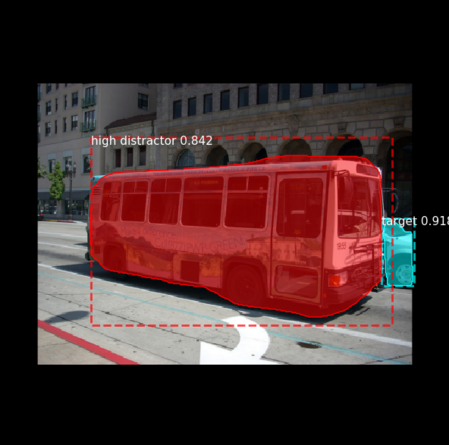}
  \label{fig:fig6}
}
\subfigure{
  \centering
  \includegraphics[scale = 0.45]{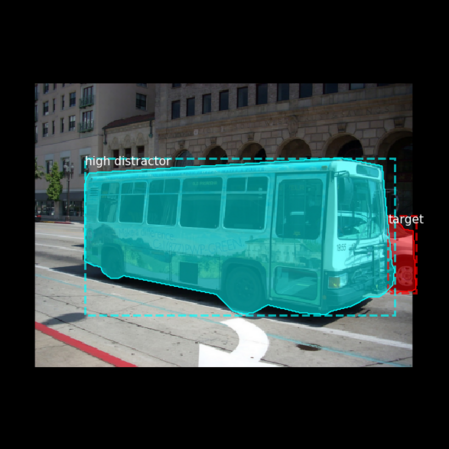}
  \label{fig:fig5}
}


\caption{Sample results of Mask-RCNN with 3 categories: target, low-distractor, and high-distractor. Left column are predictions and right column are ground-truths. The target categories from top to bottom images are bottle, bowl, and car respectively.}

\label{fig:3_class_category}
\end{figure*}

We discussed earlier that we labeled each segmented non-target object with its distraction level, i.e. the number of observers who fixated on that object. To test if a M-RCNN can distinguish distractors with different distraction levels, we trained a M-RCNN with 3 segmentation classes, corresponding to `target', `low-distractor', and `high-distractor'. Low-distractors are non-target objects fixated by 1 or 2 people, while high-distractors are those fixated by 3 or more observers. The confusion matrices and computed metrics for one run of 5 fold cross validation can be seen in figure \ref{fig:cm_3class} and table \ref{table:3class_tab}.
In general, this configuration causes a drop in the performance of the model. The MAPs decrease by 0.1-0.15, MARs decrease by 0.2-0.25, and F1-scores decrease by 0.09-0.19. Also, the network does not learn to detect low distractors. The accuracy of this class for car category is 0\%, for bowl is 1.97\%, and for bottle is 8.61\%. High-distractors are learned better by the model. The accuracy of high-distractor class for car (46\%) and bowl (43\%) categories is higher than the bottle category (23.16 \%). The target detection accuracy has also decreased a little compared to the previous configuration.
Furthermore, the networks tends to miss more low-distractors and predict them as background compared to high-distractors. For bottle, 78\% low distractors (191 out of 224) and 64\% of high-distractors (114 out of 177) are missed. For bowl, 76\% of low distractors (116 out of 152) and 0.52\% of high-distractors (98 out of 190) are missed. For car, 0.71\% (48 out of 64) of low-distractors and 0.44\% of high distractors (38 out of 87) are missed. The network also confuses some low-distractors with high-distractors to a greater degree than vice versa. Moreover, the network often considers bigger objects as high distractors and smaller ones as low-distractors. This behavior is reasonable from a logical standpoint. In addition to a higher visibility associated with bigger objects, the bigger the area of an object is the higher the probability that a fixation lies on it. Figure \ref{fig:3_class_category} shows some example results of the network in this configuration.

\begin{table*}[htbp]
\captionsetup{justification=centering}
\caption{The average performance of 3-class classification composed of low-distractor, high-distractor, and target. Results are calculated for one run of 5-fold cross validation (5 rounds in total).}

\makebox[\textwidth][c]{
    \begin{tabular}{ccccccc}
\toprule
\multirow{2}{*}{\textbf{Category}} & \multirow{2}{*}{\textbf{MAP\textsubscript{50}}}& \multirow{2}{*}{\textbf{MAR\textsubscript{50}}} & \multirow{2}{*}{\textbf{F1-score}}&
\multicolumn{3}{c}{\textbf{Accuracy}}  \\
\cmidrule(lr){5-7}

& & & & \textbf{Target} & \textbf{Low-distractor} & \textbf{High-distractor}\\
\midrule

\textbf{Bottle} & 0.35& 0.48& 0.41& 75.00\% & 8.61\% & 23.16\% \\
\midrule

\textbf{Bowl} & 0.38& 0.51& 0.43& 66.35\% & 1.97\% & 43.16\% \\
\midrule

\textbf{Car} &0.54 & 0.67& 0.60& 83.75\% & 0.00\% & 45.98\% \\
\midrule

\end{tabular}
}
\label{table:3class_tab}
\end{table*}

To better understand the impact of our two-stage fine-tuning, which is composed of first fine tuning head layers and then all layers, we compare its performance against one stage training. We consider two variations, in one variation we only fine tune the head layers and in another we fine-tune all layers of Mask-RCNN network. Table \ref{table:stage_t} summarizes our results. 
In general, training only heads of the network has a higher performance compared to training all layers in one-stage fine-tuning. Only the target detection accuracy is slightly higher when fine-tuning all layers in one stage. We hypothesise that the features that are extracted for object-class instance segmentation on COCO dataset are useful for target-distractor segmentation on COCOSearch18, thus freezing those layers and only training the heads outperforms training all layers. Two-stage tuning still outperforms both one-stage versions on all computed metric. Although one-stage tuning takes less time and occupies less memory, the performance gain from two-stage tuning outweighs the costs. 

\begin{table*}[htbp]
\captionsetup{justification=centering}
\caption{Comparison between one-stage and two stage fine-tuning. The values are averaged over 3 independent runs of 5-fold cross validation.}

\makebox[\textwidth][c]{
    \begin{tabular}{cccccccc}
\toprule
\multirow{2}{*}{\textbf{Category}} & \multicolumn{2}{c}{\multirow{2}{*}{\textbf{Fine-tune}}}& \multirow{2}{*}{\textbf{MAP\textsubscript{50}}}& \multirow{2}{*}{\textbf{MAR\textsubscript{50}}} & \multirow{2}{*}{\textbf{F1-score}}&
\multicolumn{2}{c}{\textbf{Accuracy}}  \\
\cmidrule(lr){7-8}

& & & & & &\textbf{Target} & \textbf{Distractor}\\
\midrule
\multirow{6}{*}{\textbf{Bottle}} & \multirow{2}{*}{\textbf{1-stage heads}}& \textbf{$\mu$} &0.482&	0.688&	0.567 & 73.16\% & 53.56\%\\
&  & \textbf{$\sigma$} & 0.013	& 0.007 &	0.012 &3.48\% &1.38\%\\

\cmidrule(lr){2-8}
 & \multirow{2}{*}{\textbf{1-stage all}}& \textbf{$\mu$} &0.463&	0.671&	0.548 & 75.04\% & 49.72\%\\
&  & \textbf{$\sigma$} & 0.022	& 0.016 &	0.017 &2.861\% &0.069\%\\

\cmidrule(lr){2-8}
 & \multirow{2}{*}{\textbf{2-stage heads+all}}& \textbf{$\mu$} &0.513&	0.724&	0.600 & 76.98\% & 56.32\%\\
&  & \textbf{$\sigma$} & 0.007	& 0.015 &	0.009 &4.08\% &1.28\%\\

\midrule

\end{tabular}
}
\label{table:stage_t}
\end{table*}


We compare the performance of Resnet50 backbone vs. Resnet101 for bottle category in table \ref{table:res50}. Resnet101 performs almost $\times2$ better on all computed metrics.

\begin{table*}[htbp]
\captionsetup{justification=centering}
\caption{Comparison between Resnet50 and Resnet101 backbone networks for target-distractor segmentation. The values are averaged over 5 independent runs of 5-fold cross validation.}

\makebox[\textwidth][c]{
    \begin{tabular}{cccccccc}
\toprule
\multirow{2}{*}{\textbf{Category}} & \multicolumn{2}{c}{\multirow{2}{*}{\textbf{Backbone}}}& \multirow{2}{*}{\textbf{MAP\textsubscript{50}}}& \multirow{2}{*}{\textbf{MAR\textsubscript{50}}} & \multirow{2}{*}{\textbf{F1-score}}&
\multicolumn{2}{c}{\textbf{Accuracy}}  \\
\cmidrule(lr){7-8}

& & & & & &\textbf{Target} & \textbf{Distractor}\\
\midrule

\multirow{4}{*}{\textbf{Bottle}} & \multirow{2}{*}{\textbf{Resnet50}}& \textbf{$\mu$} &0.283&	0.397&	0.330 & 38.97 \% & 29.96\%\\
&  & \textbf{$\sigma$} & 0.015	& 0.037 &	0.023 &3.50\% &2.59\%\\

\cmidrule(lr){2-8}
 & \multirow{2}{*}{\textbf{Resnet101}}& \textbf{$\mu$} &0.513&	0.724&	0.600 & 76.98\% & 56.32\%\\
&  & \textbf{$\sigma$} & 0.007	& 0.015 &	0.009 &4.08\% &1.28\%\\

\midrule

\end{tabular}
}
\label{table:res50}
\end{table*}


To understand how augmenting the dataset with horizontal and vertical flipping affects the performance of the model, we compared model performance in 4 cases: 1-no augmentation, 2-just augmenting the first stage 3-just augmenting the second stage 4-augmenting both stages (the default case). The results are listed in table \ref{table:aug}. Adding augmentation improves the results on all metrics. Also, the improvement achieved by adding augmentation to the second stage is higher than augmenting the first stage. The reason could be that the second stage has a higher architecture complexity as we train all M-RCNN layers, compared to the first stage where we only train the heads. Augmentation is a good strategy to avoid overfitting when training larger networks with small datasets. The computed metrics for augmenting only the second stage and augmenting both stages are quite similar, with slightly higher variance in the results of different runs for the first case. 

\begin{table*}[htbp]
\captionsetup{justification=centering}
\caption{The impact of augmentation on different training stages. The values are averaged over 3 independent runs of 5-fold cross validation.}

\makebox[\textwidth][c]{
    \begin{tabular}{cccccccc}
\toprule
\multirow{2}{*}{\textbf{Category}} & \multicolumn{2}{c}{\multirow{2}{*}{\textbf{Augmentation}}}& \multirow{2}{*}{\textbf{MAP\textsubscript{50}}}& \multirow{2}{*}{\textbf{MAR\textsubscript{50}}} & \multirow{2}{*}{\textbf{F1-score}}&
\multicolumn{2}{c}{\textbf{Accuracy}}  \\
\cmidrule(lr){7-8}

& & & & & &\textbf{Target} & \textbf{Distractor}\\
\midrule
\multirow{8}{*}{\textbf{Bottle}} & \multirow{2}{*}{\textbf{No augmentation}}& \textbf{$\mu$} &0.509&	0.707&	0.592 & 74.14\% & 53.82\%\\
&  & \textbf{$\sigma$} & 0.003	& 0.009 &	0.005 &5.17\% &0.67\%\\

\cmidrule(lr){2-8}

 & \multirow{2}{*}{\textbf{1st stage augmented}}& \textbf{$\mu$} &0.513&	0.712&	0.597 & 76.91\% & 54.27\%\\
&  & \textbf{$\sigma$} & 0.002	& 0.017 &	0.008 &1.14\% &3.02\%\\

\cmidrule(lr){2-8}
 & \multirow{2}{*}{\textbf{2nd stage augmented}}& \textbf{$\mu$} &0.525	& 0.722&	0.607 & 77.18\% & 55.21\%\\
&  & \textbf{$\sigma$} & 0.021	& 0.021 &	0.020 &4.04\% &2.31\%\\

\cmidrule(lr){2-8}
 & \multirow{2}{*}{\textbf{Both stages augmented}}& \textbf{$\mu$} &0.513&	0.724&	0.600 & 76.98\% & 56.32\%\\
&  & \textbf{$\sigma$} & 0.007	& 0.015 &	0.009 &4.08\% &1.28\%\\

\midrule

\end{tabular}
}
\label{table:aug}
\end{table*}





\subsection{Discussion}

The method we discussed here, with its relatively quick training and inference, is a useful way of segmenting target and distracting objects in images during visual search. 

One drawback of our method is that we need to train a separate Mask-RCNN for each target category. One possible future direction is to unify these category-specific models into one model. 

Another shortcoming of our method is that we only consider an object as fixated if the fixation location lies exactly on its segmentation. However, it might be the case that multiple objects are fixated by each fixation. This would need us to consider a radius (similar to the high-resolution fovea image which typically covers a region of  1.5° of the visual field) around each fixation location and consider all objects whose segmentation partially lies under this radius as fixated. 

Our network often predicts bigger objects as more distracting. This behavior might be caused by the higher visibility of larger objects or our current configuration, as more fixations lie on the area of larger objects. However, if we consider a radius around the fixations, then at each fixation many of the smaller nearby objects are also attended and might be considered as distracting as the bigger object. 

Furthermore, object-based visual attention has the assumption that all parts of an object are equally salient, or in our case equally distracting. This assumption often does not hold true. For instance, in one of the search images there is a big monitor displaying images that attract most observers attention. While these images are only placed on the monitor's screen, we consider the whole monitor as highly distracting.

It is worth noting that visual search is similar to object detection task in computer vision. In fact in both of our methods, we employ pre-trained networks for object-detection task to predict visual search behavior of humans. State-of-the-art object detectors such as Mask-RCNN (and faster-RCNN) have object proposal networks that suggest the most probable regions that could contain an object. Similarly, when searching for an object, humans fixate at locations that are likely to contain the target object. The contextual information/gist (which is assumed to be computed pre-attentively), and prior knowledge about where an object should be located affect these distracting locations/objects. 

In terms of internal architecture, a CNN-based object detector and visual attention theories share some similarities. For instance, CNN uses kernels (filters) each corresponding to a specific feature. Each filter attempts to find the locations in the image where the activation of that feature is maximized. Then during network training, the best filters (features) for detecting a specific target object are chosen. Feature integration theory \cite{TREISMAN198097} also suggests that in conjunction search, humans scan the scene in a serial way, merging different features of each fixated location until finding the target. Feature integration theory proposes that human brain possesses feature-specific neurons that activate in locations where those features are present. However, unlike a CNN where features are calculated across the whole image pixels; humans only perceive features around the fixated location in an image. This is due to the fact that humans' eye receive high-resolution image in a small region within the fovea; hence, eye movements are needed to perceive different parts of the scene with high resolution. The other reason is the computational limitation of human brain, which makes it difficult to perceive the whole scene with high-resolution. 

A difference our segmentation model has with RCNN object detectors, besides using fixated objects for training the network, is that it looks for two object classes: target and distractor. The target object can only be one object category, for example in case of searching for bottle, only bottles can be the target objects. However, any other objects categories could be considered as distractors, for instance: glass, chair, spoon, fork, etc. While, an RCNN aims at segmenting multiple object classes which correspond to different object categories.

\section{Conclusions}

In summary, we introduced two deep-learning based approaches to model human visual attention and distraction behavior during visual search tasks. Although our models share similarities to object detectors in computer vision, the main purpose of our work is not only detecting the target but also the distractors, to explore how human observers search an image for a particular target. 

Both of our methods provide evidence that distracting regions or objects in images during visual search are predictable. Our first model's predicted fixation maps show that the model learns where to look for each target category. For instance, when looking for oven or toilet, the locations closer to the ground were considered salient; whereas for TV or clock the higher items in images were often considered salient. Moreover, the similarity of the items to the target led to higher distraction. The items sharing some features with the objects of target category were sometimes considered salient both by the network and human observers.

Our second method involves training separate Mask-RCNN networks for each target category to segment and classify the target and distracting objects. This network outputs a confidence score for each classified object, and by defining the right confidence threshold for each image, we were able to detect the most distracting objects existing in that image. We interpreted that several factors affect the level of distraction such as the size of the distractor, object class of the distractor, similarity to the target, proximity to the target, and proximity to the center of the image. 

A potential application of our methods could be in visual marketing, i.e. using visual information to guide customers' attention. Our methods could be extended to 3D environments and be applied to supermarkets for analyzing customers behavior. Learning how customers are distracted to different products can help us to guide customers attention toward more healthy products. For application to a supermarket, one might extend our approach to predict visual behavior during visual foraging tasks, where multiple target categories are being searched at the same time. This would require collecting a visual foraging dataset.

One limitation of modeling task-oriented visual attention is the scarcity of fixation-labeled datasets. This problem might be due to the difficulty associated with collecting eye tracking data, especially with an added visual task. COCOSearch18 is the first dataset that provides fixation labels for searching a relatively large number of target categories (18). Even though COCOSearch18 has good potential for conducting further behavioral research on human's visual attention, it only contains the fixation locations of 10 observers. We believe that for a more reliable modeling of distraction, a greater number of observers are needed, so that the individual preferences are less prominent in the modeling. Currently, our models are trained and tested on the fixation data of these 10 observers, who might not be a good representative of a larger population. 

Another possible direction for future improvement could be to merge our first and second method, such that the predictions of the fixations density maps are transformed into object-based segmentation. There exists multiple methods that propose object segmentation based on fixation points such as \cite{5459254}, which we did not explore due to time constraint.

All in all, both of our proposed methods are good starting points in modeling and understanding human's visual search behavior. Further research is needed to validate our results, and develop stronger predictors for humans' distraction during visual search.

\section{Acknowledgments}

We would like to thank Institute for Data Valorization (IVADO) and Natural Sciences and Engineering Research Council of Canada (NSERC) for providing funding for this work.

Commercial relationships: None.
Corresponding author: Manoosh Samiei.
Email: manoosh.samiei@mail.mcgill.ca.
Address: Center for Intelligent Machines, Department
of Electrical and Computer Engineering, McGill
University, Montreal, Quebec, Canada.


\bibliography{Predicting_Visual_Attention_Distraction.bib}{}
\bibliographystyle{IEEEtran}

\section{Appendix}

\end{document}